\documentclass{article} 
\usepackage{collas2026_conference,times}
\usepackage{easyReview}

\usepackage{amsmath,amsfonts,bm}

\def\eqref#1{equation~\ref{#1}}

\def\1{\bm{1}}

\DeclareMathAlphabet{\mathsfit}{\encodingdefault}{\sfdefault}{m}{sl}
\SetMathAlphabet{\mathsfit}{bold}{\encodingdefault}{\sfdefault}{bx}{n}

\usepackage{hyperref}
\hypersetup{
    colorlinks=true,
    linkcolor=red,
    filecolor=magenta,
    urlcolor=blue,
    citecolor=purple,
    pdftitle={Overleaf Example},
    pdfpagemode=FullScreen,
    }

\usepackage[ruled,vlined,linesnumbered]{algorithm2e}
\SetAlgoNlRelativeSize{0}

\usepackage{amssymb}
\usepackage{tikz}
\usepackage{siunitx}
\usepackage{float}
\usepackage{tabularx}   
\usepackage{multirow}
\usepackage{caption}    
\usepackage{csquotes}

\usepackage{enumitem} 

\usepackage{array}
\usepackage{booktabs}
\usepackage{wrapfig}

\usepackage{amsmath}
\collasfinalcopy

\title{Beyond Single-Model Optimization: Preserving Plasticity in Continual Reinforcement Learning}  

\author{Lute Lillo, Nick Cheney \\
Department of Computer Science\\
University of Vermont\\
USA \\
\texttt{\{elillopo, ncheney\}@uvm.edu} \\
}

\preprintcopy 

\begin{document}

\maketitle

\begin{abstract}
Continual reinforcement learning must balance retention with adaptation, yet many methods still rely on \emph{single-model preservation}, committing to one evolving policy as the main reusable solution across tasks. 
Even when a previously successful policy is retained, it may no longer provide a reliable starting point for rapid adaptation after interference, reflecting a form of \emph{loss of plasticity} that single-policy preservation cannot address.
Inspired by quality-diversity methods, we introduce \textsc{TeLAPA} (Transfer-Enabled Latent-Aligned Policy Archives), a continual RL framework that organizes behaviorally diverse policy neighborhoods into per-task archives and maintains a shared latent space so that archived policies remain comparable and reusable under non-stationary drift.
This perspective shifts continual RL from retaining isolated solutions to maintaining \emph{skill-aligned neighborhoods} with competent and behaviorally related policies that support future relearning.
In our MiniGrid CL setting, \textsc{TeLAPA} learns more tasks successfully, recovers competence faster on revisited tasks after interference, and retains higher performance across a sequence of tasks.
Our analyses show that source-optimal policies are often not transfer-optimal, even within a local competent neighborhood, and that effective reuse depends on retaining and selecting among multiple nearby alternatives rather than collapsing them to one representative.
Together, these results reframe continual RL around reusable and competent policy neighborhoods, providing a route beyond single-model preservation toward more plastic lifelong agents.
\end{abstract}

\vspace{-0.5em}
\section{Introduction}
\label{sec:intro}
\vspace{-0.5em}
Training neural networks on a sequence of tasks exposes a fundamental stability--plasticity dilemma. This tension leads to two primary failure modes: \textit{Catastrophic Forgetting}, where a network's performance on previously learned tasks degrades as it acquires new information \cite{mccloskey1989catastrophic}, and \textit{Loss of Plasticity}, where the agent's ability to learn \emph{new} tasks diminishes after continuous training \citep{abbas2023loss,dohare2024loss}. While forgetting is often framed as overwriting weights important to older tasks \citep{kirkpatrick2017overcoming}, loss of plasticity reflects a degradation of the learning mechanism itself, leading to slow convergence or an inability to adapt even when retention is otherwise strong \citep{lyle2024disentangling,lyle2022understanding}.

Most continual learning methods operate within a \emph{single-model} paradigm: they preserve one evolving parameter vector by regularizing updates \citep{kirkpatrick2017overcoming},
replaying past data \citep{rebuffi2017icarl}, or allocating additional parameters via growth/routing/pruning \citep{mallya2018packnet}. Despite their differences, these approaches are predominantly trapped in a \textbf{single-model preservation paradigm}. They protect or modify a single converged solution—a sharp, highly specialized point in parameter space that can be unstable under distribution shift and interference. Consequently, preserving a single solution can inadvertently undermine the capacity for future adaptation. \citep{abbas2023loss,kumar2023maintaining,nikishin2022primacy,lewandowski2023directions,dohare2021continual}.

We instead ask a slightly different question: under non-stationarity and interference, what must be preserved so that the agent remains able to learn well \emph{later}? 
Our premise is that future learnability is often not a property of one parameter vector, but of a \emph{recoverable neighborhood} around a competent solution. 
A single policy may retain immediate competence yet fail to support later adaptation, whereas a neighborhood of behaviorally related policies can preserve multiple nearby starting points for future learning.

Therefore, we propose that the relevant object of preservation in continual RL is not a point solution, but a \emph{skill neighborhood}: a local manifold of behaviorally distinct policies. 
Inspired by Quality-Diversity (QD) optimization \citep{mouret2015illuminating,pugh2016quality,lehman2011novelty}, we introduce \textbf{TeLAPA}\footnote{Code available at: \url{https://github.com/lute47lillo/telapa_collas2026}} (\emph{Transfer-Enabled Latent-Aligned Policy Archives\footnote{TeLAPA is named after te lapa, a Polynesian wayfinding phenomenon described as faint flashing light on the ocean used to orient toward land; analogously, our latent policy archives serve as an illumination-based navigational aid in policy space}}). 
\textsc{TeLAPA} trains a base policy with PPO \citep{schulman2017proximal} and populates an archive of diverse elites around it using parameter-space MAP-Elites \citep{mouret2015illuminating}. 
During continual transfer, candidate initializations are retrieved from prior archives using latent-space diversity and then selected through short transfer-time evaluations.
This archive-then-retrieve instrument allows us to evaluate continual learning in terms of \emph{latent-space geometry}, answering not just whether a prior policy was retained, but whether a useful neighborhood remains \emph{navigable}.

\textsc{TeLAPA} relies on a learned trajectory embedder to place archived policies into a shared latent behavior space, where nearby policies are intended to reflect behaviorally similar solutions.
However, because learned representations are non-stationary, the latent coordinates of archived behaviors can drift and scramble these neighborhood relations, degrading retrieval and downstream reuse even with replay \citep{caccia2021new, lin2023pcr}. 
Similar alignment challenges exist in unsupervised QD \citep{grillotti2022unsupervised, olle2021pga}. Thus, maintaining a shared latent geometry under encoder drift becomes central to the continual-learning problem itself.

\vspace{-0.5em}
\paragraph{Contributions \& Roadmap.}
We reframe continual RL around \emph{latent manifold dynamics}, advancing two main claims: (i) transfer should be evaluated by the preservation of useful behavioral neighborhoods rather than isolated policies, and (ii) stabilizing the embedding coordinate system is essential for retrieval. To achieve this, \textsc{TeLAPA} employs a robust embedding procedure using \emph{anchor sets, replay/alignment losses}, and \emph{periodic re-embedding} to mitigate drift across non-stationary curricula.

Section~\ref{sec:related} contextualizes our work within continual plasticity, representation drift, and Quality-Diversity (QD) behavior embeddings.
Section~\ref{sec:method} formalizes continual RL with revisits, introducing \textsc{TeLAPA} as a neighborhood-preserving alternative to single-model retention.
Section~\ref{sec:v2} details our maintenance procedure---utilizing anchor banks, replay updates, normalizer refitting, and periodic re-embedding---to ensure the shared latent behavior space remains navigable under encoder drift.
Section~\ref{sec:results_baseline} evaluates whether these illuminated neighborhoods improve transfer compared to single-model baselines, and analyze whether \textsc{TeLAPA} preserves a longer-horizon stepping-stones structure.

\vspace{-1.0em}
\section{Related Work}
\label{sec:related}

\vspace{-0.6em}
This section surveys prior work relevant to our geometry-centric view of continual RL, organized around:
(i) mechanisms of plasticity loss and representational pathologies,
(ii) representation drift and encoder alignment under non-stationarity,
(iii) quality-diversity and unsupervised behavior embeddings, and
(iv) policy-manifold and basin-geometry perspectives.

Together, these four lines of work motivate our framework. CL methods based on regularization, replay, or routing expose the limits of preserving a single model; works on representation-drift show that feature-space neighborhoods can become unstable under non-stationarity; QD methods provide a mechanism for preserving local repertoires rather than one optimum; and policy-manifold geometry suggests that transfer depends on basin structure, not only isolated solutions. 
Our contribution is to connect these ideas in continual RL to preserve \emph{skill neighborhoods}, organize them in a shared latent behavior space, and study both their utility and their failure modes under representation drift.

\vspace{-0.6em}
\paragraph{Beyond the single-model plasticity bottleneck.}
Most continual learning approaches---whether utilizing regularization \citep{li2017learning, kirkpatrick2017overcoming, chung2024parseval}, replay \citep{rebuffi2017icarl, chaudhry2018efficient, rolnick2019experience, lin2023pcr}, architectural expansion \citep{mallya2018packnet, yoon2017lifelong, yang2021grown}, or weight resets \citep{frati2023reset, kumar2023maintaining}---operate within a \emph{single-model} paradigm. 
We argue this focus on preserving a single, highly specialized policy exacerbates the plasticity bottleneck \citep{klein2024plasticity}. 
Converged RL solutions are inherently fragile to interference and lack the adaptability required for future tasks, often performing worse than random initialization. 
This loss of plasticity is the primary hurdle for long-term autonomy and stems from a complex intersection of representational issues. 
While researchers have identified contributing factors such as dead or dormant units \citep{maas2013rectifier, sokar2023dormant}, impact of activation functions \citep{delfosse2021adaptive, lewandowskiplastic, lillo2025activation}, reduced gradients \citep{abbas2023loss}, increasing parameter norms \citep{nikishin2022primacy}, rank-deficient curvature \citep{lyle2022understanding,lewandowski2023directions}, and declining diversity \citep{kumar2020implicit,kumar2023maintaining,dohare2024loss}, no single mechanism fully explains its onset. 
Instead, it resembles a ``Swiss cheese'' of overlapping failures \citep{lyle2024disentangling}. 
Because single point solutions are so vulnerable to these cascading representational collapses, our approach abandons single-model preservation to instead maintain robust, adaptable \emph{solution neighborhoods}. 
Therefore, we study plasticity preservation operationally: as the learner's ability to recover and re-adapt after interference. This makes our work complementary to mechanistic diagnostics of plasticity loss; rather than isolating a single within-network cause, we ask whether changing the preserved object from one policy to a structured neighborhood improves the system's functional capacity for later adaptation.

\vspace{-0.6em}
\paragraph{Representation drift and encoder alignment under non-stationarity.}
In continual learning, the learned embedding itself can drift as training proceeds, so representations of previously seen data (and thus neighborhood relations) change under new-task updates, degrading retrieval and transfer even when replay is present \citep{caccia2021new}. Recent methods therefore treat stability as a \emph{feature-space} problem: replay is augmented with contrastive/metric objectives that explicitly preserve neighborhood structure, e.g., proxy-based contrastive replay (PCR) built on supervised contrastive learning \citep{lin2023pcr, khosla2020supervised}. 
Correspondingly, drift can be diagnosed geometrically rather than only through task accuracy, via prototype/centroid displacement and neighborhood-graph overlap measures such as $k$NN graph similarity (Jaccard-style agreement), which directly quantify whether local neighborhoods are preserved across time \citep{tavares2024measuring}. 
These tools are directly relevant to continual RL with latent behavior embeddings, where policy \emph{vicinity} is the substrate for archive retrieval and where non-stationary encoders can otherwise invalidate cross-time comparisons.

\vspace{-0.6em}
\paragraph{Quality-diversity and unsupervised behavior embeddings.}
Quality-Diversity (QD) algorithms, such as MAP-Elites \citep{mouret2015illuminating}, aim to generate a diverse archive of high-performing solutions rather than a single optimum \citep{pugh2016quality}. Classical MAP-Elites relies on handcrafted behavior descriptors, while AURORA introduced \emph{learned} behavior embeddings (autoencoder latents on trajectories) as descriptors, enabling unsupervised discovery of behavioral structure \citep{grillotti2022unsupervised}. Recent work combines QD with gradient-based updates and learned descriptors (e.g., PGA-MAP-Elites and PGA-AURORA) \citep{olle2021pga,chalumeau2022neuroevolution}, while other extensions have explored multi-task and parameterized settings \citep{mouret2020quality, anne2023multi, anne2024parametric}, they typically assume tasks are known or optimized simultaneously. In parallel, unsupervised skill discovery in RL (e.g., DIAYN) learns latent skill variables that induce diverse behaviors and can be viewed as a complementary route to behavior embedding \citep{eysenbach2018diversity}. In contrast, CL faces sequential tasks under interference, where the key challenge is not only to preserve past competence but to preserve \emph{future learnability}.
In contrast, we adapt the archive-based view of QD to continual RL by preserving skill neighborhoods—sets of competent, behaviorally related policies—rather than a single retained policy per task.

\vspace{-0.6em}
\paragraph{Policy manifold and basin geometry viewpoints.}
A growing literature studies the geometry of solution landscapes, showing that many distinct minima can be connected by low-loss curves (mode connectivity) \citep{garipov2018loss,draxler2018essentially,frankle2020linear}.
\cite{rahn2023policy} studies return landscapes in continuous control and shows that policies from the same training run can often be connected by simple parameter-space paths without low-return barriers, while also revealing that local neighborhoods around policies can differ substantially in post-update return stability.
In RL and multi-task RL, connectivity and policy-subspace approaches suggest that task solutions may share basins or lie on lower-dimensional manifolds that can be exploited for transfer \citep{hendawy2025allconnected,gaya2022building,rakicevic2021policy}. 
These perspectives motivate our focus on transfer-relevant policy geometry in continual RL, viewed through a learned latent behavior space. 
In particular, we ask not only whether solutions are connected, but whether they retain a recoverable neighborhood of behaviorally related policies that can support later adaptation.

\vspace{-0.5em}
\section{Framework: Policy Archives for Continual RL}
\label{sec:method}
\vspace{-0.8em}
\subsection{Continual RL setting with revisits}
\vspace{-0.5em}
Inspired by other recent experimental settings for continual learning \citep{abbas2023loss} we designed our evaluation such that the environment cycles over through a fixed sequence of tasks $\mathcal{S} = (T_1,\dots,T_K)$, where each $T_k$ is an MDP and a policy $\pi_\theta$ is trained to maximize the expected discounted return on the current task. 
Our focus is not only \emph{retention} of past competence, but \emph{functional plasticity preservation}: the ability to rapidly (re)adapt after interference from intervening tasks and the transferability of skills between tasks. 
Therefore, we evaluate plasticity at the system level, through recoverability and transfer from stored policy neighborhoods, rather than through a single-model diagnostics.

\vspace{-0.6em}
\paragraph{MiniGrid instantiation.}
In our experiments, tasks $T_k$ are goal-oriented MiniGrid environments \citep{MinigridMiniworld23}:
partially observable grid-worlds in which an agent receives a mission specification together with a local symbolic/image observation, and must navigate and interact with objects using a small discrete action space under sparse success-based rewards.
To isolate plasticity under interference, we construct a primary curriculum $\{A,B,C,D,E\}$ \citep{narvekar2020curriculum}, and corresponding revisits $\{A',B',\dots\}$ by selecting environments that progressively increase interaction and long-horizon structure, while keeping evaluation comparable via consistent success-rate metrics and fresh seeds per visit. 
In these environments success is obtained only when the agent completes the required goal. 
Thus, the curriculum is not only a sequence of different layouts, but a sequence of reusable interaction primitives: navigation and text grounding, object pickup, key--door unlocking, clutter manipulation, and longer-horizon subgoal chaining.
In App.~\ref{app:curriculum_ablation} and App.~\ref{app:long_curriculum} we evaluate different curriculum configurations to avoid designing only favorable hand-crafted sequences.
In this setting, revisit performance primarily reflects whether the learner can \emph{re-enter} a suitable behavioral basin after interference, rather than whether it can discover the task from scratch (see App.~\ref{app:env_revisits} for deeper discussion).

\vspace{-0.5em}
\subsection{\textsc{TeLAPA}: Archive-then-retrieve policy neighborhoods}
Rather than preserving a single parameter vector across the stream, \textsc{TeLAPA} maintains a \emph{library of per-task policy neighborhoods}.
For each task $T_k$, we first train a competent base policy $\theta^{\mathrm{base}}_{T_k}$ with PPO \citep{schulman2017proximal}.
Then, we run a post-training illumination phase around this base policy: mutated offspring $\theta'=\theta+\delta$ are evaluated on $T_k$, assigned a source-task fitness $f_{T_k}(\theta')$, and embedded into a shared latent behavior space.
The illumination step does not assume that arbitrary weight perturbations are useful.
The resulting archive
$\mathcal{A}_{T_k}$ stores $N_k$ competent but behaviorally diverse elite entries for task $T_k$, where $N_k = |\mathcal{A}_{T_k}|$ denotes the realized archive size after illumination.
In all experiments, the illumination archive used a target size of $256$ elites and a maximum capacity of $384$ elites, so $N_k \leq 384$ and is kept close to $256$ by the adaptive archive-spacing threshold.
We denote the $i$-th elite entry in archive $\mathcal{A}_{T_k}$ by $\eta_{k,i}$.
Across tasks, the archive library $\{\mathcal{A}_{T_m}\}_{m\le k}$ serves as a set of candidate basins that can later be navigated when adaptation is required on a revisit or new task. 

\vspace{-0.5em}
\paragraph{Candidate pooling.}
Rather than restricting transfer to the immediately preceding task, \textsc{TeLAPA} builds the candidate pool from the union of all previously learned task archives, so that any prior archive can serve as a source of transfer candidates.
First, when a new task $T_{k+1}$ arrives, we retrieve an initialization from the library of previously learned task archives: $\mathcal{U}_{k+1}=\bigcup_{m \le k} \mathcal{A}_{T_m}$.
Then, we construct a compact candidate pool 
$\mathcal{C}_{k+1}=\{\eta_j\}_{j=1}^{K_{\mathrm{pool}}}\subseteq \mathcal{U}_{k+1}$ using a diversity-aware down-selection procedure in the shared latent space.
We implement this down-selection as greedy farthest-point selection in the globally normalized latent space (see App.~\ref{app:global_normalizer}). 
Starting from a high source-task fitness elite, $\eta_1=\arg\max_{\eta\in\mathcal{U}_{k+1}} f_{\eta}$, each subsequent candidate is chosen to maximize its minimum latent distance to the already selected candidates until $K_{\mathrm{pool}}$ candidates are selected.
The candidate pool covers distinct behavioral regions of the prior archive library rather than many near-duplicates of the same source solution.
Thus, few-shot evaluation explores multiple plausible transfer origins rather than minor variations of the same source solution.

\vspace{-0.5em}
\paragraph{Few-shot origin selection.}
For each candidate elite $\eta_j \in \mathcal{C}_{k+1}$ on the incoming task $T_{k+1}$, we measure zero-shot performance and run a short adaptation probe, resetting optimizer state each time so that the probe reflects basin recoverability rather than optimizer momentum.
Selection is conservative: we first keep only candidates whose final probe performance is close to the best observed value, and among those near-best candidates we prefer the one showing the strongest short-horizon recoverability. 
The selected initialization is designed to explicitly target \emph{recoverable plasticity}: it is the source elite whose basin appears most readily reusable for the new task, not necessarily the one with the highest few-shot score.

\vspace{-0.5em}
\subsection{Preserving the Latent Space Under Drift}
\label{sec:v2}

In continual RL, interference affects not only policy parameters, but also the representation used to organize archived behaviors. 
In \textsc{TeLAPA}, retrieval depends on a shared latent space to assess and maintain behavioral diversity, so archived policies remain useful only as long as latent proximity continues to reflect comparable behavior over time. 
As new tasks arrive, updates to the trajectory embedder and its normalization can shift previously stored behaviors in latent space, changing neighborhood relations even when the archived policies themselves remain unchanged. 
An archive may therefore stay populated while becoming progressively harder to navigate and reuse.
 
Therefore, \emph{we treat latent-space stability as part of the continual-learning problem}. 
Preserving an archive means preserving not only its policies, but also the neighborhood structure that makes those policies retrievable for future adaptation. 
Thus, the learned trajectory embedding is used as an archive coordinate system, not as an additional policy input or intrinsic reward.
Our goal is not to freeze the representation, but to maintain a latent geometry that can evolve without destroying the behavioral organization needed to make stored policies reusable.
Inspired by QD methods with stationary re-embedding procedures such as AURORA \citep{grillotti2022unsupervised}, we augment \textsc{TeLAPA} with three coupled mechanisms: \emph{anchor sets}, \emph{replay-based alignment}, and \emph{periodic re-embedding of archives}.

\vspace{-0.5em}
\paragraph{Archive Representation Under Drift.}
Let $\phi_t$ denote the shared trajectory embedder after task boundary $t$, and let $\mathcal{A}_k$ denote the archive for task $T_k$. 
Each elite ($\eta_{k,i}$) stores its policy parameters ($\theta_{k,i}$), latent descriptor ($z_{k,i}$), and source-task fitness ($f_{k,i}$) by computing it all from its rollout trajectory $\tau_{k,i}$. 
Specifically, let $\Xi_{k,i}=\{\tau_{k,i}^{(1)},\ldots,\tau_{k,i}^{(M)}\}$ denote the $M$ evaluation episodes collected by running policy $\pi_{\theta_{k,i}}$ on $T_k$, and let $R(\tau_{k,i}^{(m)})$ be the episodic return of episode $m$. 
Then, 
\begin{equation}
\mathcal{A}_k = \left\{(\theta_{k,i}, f_{k,i}, z_{k,i}^{(t)})\right\}_{i=1}^{N_k},
\qquad
z_{k,i}^{(t)} = \phi_t(\tau_{k,i}).
\qquad
f_{k,i}
=
\frac{1}{M}\sum_{m=1}^{M} R\!\left(\tau_{k,i}^{(m)}\right).
\label{eq:archive}
\end{equation}
When $\phi_t$ changes, previously stored descriptors may no longer be expressed in the current coordinate system, degrading retrieval even if the underlying archived policies are unchanged. 
This drift matters because these descriptors are not passive summaries of past behavior; they define the archive geometry used by \textsc{TeLAPA}. 
They determine whether a competent candidate is behaviorally distinct enough to be inserted, support latent-diverse candidate pooling before few-shot origin selection, and provide the basis for diagnostic analyses of archive geometry and lineage structure.

\vspace{-0.5em}
\paragraph{Anchor Sets and Replay-Based Alignment.}
At each task boundary, we maintain two banks of past episode sets: an \emph{anchor} bank of reference behaviors and a \emph{replay} bank of broader episode sets. 
We freeze a ``teacher'' copy of the current embedder and update a live ``student'' embedder on mixed batches drawn from new-task data and both banks. 
The student is trained with a composite objective that makes the latent space informative while limiting arbitrary drift across task boundaries. 
We give the compact form here and provide the full batch construction, augmentation procedure, and implementation details in App.~\ref{app:teacher_student_update}--\ref{app:total_emb_loss}:
\begin{equation}
\mathcal{L}_{\mathrm{emb}} =
w_{\mathrm{contrast}} \mathcal{L}_{\mathrm{contrast}}
+
w_{\mathrm{distill}} \mathcal{L}_{\mathrm{distill}}
\label{eq:embed_train_obj}
\end{equation}
Let $z_i^{(1)}=\phi_t(\widetilde{\tau}_i^{(1)})$ and $z_i^{(2)}=\phi_t(\widetilde{\tau}_i^{(2)})$ denote student embeddings of two augmented views of the same sampled episode. 
We define the temperature-scaled cosine similarity logit between any of these views as $s_{ij}=\bar z_i^{(1)\top}\bar z_j^{(2)}/\tau_{\mathrm{emb}}$ where $\bar z=z/\|z\|_2$, and use a symmetric InfoNCE loss \citep{oord2018representation} as the contrastive term: 
\begin{equation}
\mathcal{L}_{\mathrm{contrast}}
=
-\frac{1}{2B}\sum_{i=1}^{B}
\left[
\log
\frac{\exp(s_{ii})}{\sum_{j=1}^{B}\exp(s_{ij})}
+
\log
\frac{\exp(s_{ii})}{\sum_{j=1}^{B}\exp(s_{ji})}
\right].
\end{equation}
The contrastive term is used to learn the comparison geometry for archived behaviors.
Each training item contributes two augmented views of the same logged episode. 
The numerator aligns the two views from the same episode, while the denominator compares that episode against all other sampled episodes in the boundary batch.
Thus, InfoNCE defines the comparison geometry used by the archive: embeddings are encouraged to remain stable under small trajectory-level augmentations, while different rollout traces remain distinguishable in the batch.
In \textsc{TeLAPA}, behavioral diversity is then measured as distance between policy-level mean descriptors in this globally normalized latent space: two elites are considered behaviorally diverse when they occupy separated regions of the learned trajectory space.

This use of InfoNCE differs from unsupervised skill-discovery objectives. 
Methods such as DIAYN \citep{eysenbach2018diversity} use diversity objectives to induce policies that visit distinguishable behaviors. 
By contrast, \textsc{TeLAPA} does not use the contrastive loss as an intrinsic reward and does not update policy parameters through this objective. 
Policy competence is determined by environment reward during PPO training and MAP-Elites insertion; InfoNCE only trains the latent space used to compare, maintain, and retrieve already-evaluated policies.

For the anchor episodes subset $\mathcal{I}_{\mathrm{anc}}$, distillation is computed on the original non-augmented episodes using the frozen teacher $\tilde{\phi}_t$ and the live student $\phi_t$:
\begin{equation}
\mathcal{L}_{\mathrm{distill}}
=
\frac{1}{|\mathcal{I}_{\mathrm{anc}}|d}
\sum_{i\in\mathcal{I}_{\mathrm{anc}}}
\left\|
\frac{\phi_t(\tau_i)}{\|\phi_t(\tau_i)\|_2}
-
\frac{\tilde{\phi}_t(\tau_i)}{\|\tilde{\phi}_t(\tau_i)\|_2}
\right\|_2^2
+
\lambda_{\mathrm{norm}}
\frac{1}{|\mathcal{I}_{\mathrm{anc}}|}
\sum_{i\in\mathcal{I}_{\mathrm{anc}}}
\left(
\|\phi_t(\tau_i)\|_2
-
\|\tilde{\phi}_t(\tau_i)\|_2
\right)^2.
\end{equation}
The distillation term reduces representation drift by keeping the student's embeddings of anchor episodes only close to those produced by the frozen teacher in both latent direction and scale, so that reference behaviors remain stable while the representation adapts to new tasks.

\vspace{-0.5em}
\paragraph{Periodic Archive Re-embedding.}
After each embedder update, we re-embed previously stored archives in the current latent space by recomputing each elite's policy-level descriptor from its stored evaluation episodes under the updated embedder and global normalizer:
$z_{k,i}^{(t+1)} = \mathcal{N}_{t+1}\!\left(\frac{1}{M}\sum_{m=1}^{M}\phi_{t+1}(\tau_{k,i}^{(m)})\right)$.
Here, $\mathcal{N}_{t+1}$ denotes the global latent normalizer refit after the boundary update; see App.~\ref{app:global_normalizer}.
This ensures that downstream retrieval and geometry analysis are performed in a consistent coordinate system.



\section{Preserving policy neighborhoods improves continual transfer}
\label{sec:results_baseline}

We first test whether preserving policy neighborhoods yields better continual transfer than preserving a single evolving model.
To evaluate transfer and forgetting, we report the \textbf{Standardized Time-To-Threshold (TTT)}, which measures recoverability by the number of environment steps required to reach a competence target performance level $\tau_T$ upon task revisit. 
For each task, this target $\tau_T$ is set to $0.9$ times the mean best success rate (SR) achieved by task-specific \textsc{Scratch}, ensuring a rigorous ``baseline-to-beat'' standard. 
Alongside TTT, we provide a diagnostic for \textbf{Backward Transfer (BWT)} to address where a model that fails to learn a task cannot technically forget it. 
This diagnostic decomposes retention into three additional metrics: 
(i) \textbf{Coverage}, the fraction of tasks that reached $\tau_T$ during initial training; 
(ii) \textbf{normalized BWT (nBWT)}, measuring the relative performance drop exclusively on previously learned tasks ($\mathrm{SR}(T)\ge\tau_T$) to isolate true forgetting from non-learning; and 
(iii) \textbf{Threshold Retention (TR)}, the fraction of tasks remaining above $\tau_T$ at the end of the sequence. 
Together, these diagnostics are not intended to identify a unique mechanistic cause of plasticity loss inside one network.
Instead they clarify whether performance is due to a failure to acquire knowledge, a failure to retain it, or an improved capacity for rapid recovery (see App.~\ref{app:metrics_transfer}). 

We compare \textsc{TeLAPA} against a diverse set of baselines, each representing a different form of single-model preservation or transfer: (i) \textbf{Scratch:} train each task independently from random initialization; (ii) \textbf{Scratch-Reuse:} train from scratch the first time a task is seen, store the resulting task-specific policy, and reuse only that policy when the same task reappears; (iii) \textbf{Finetune:} continue training a single PPO instance across the task sequence; (iv) \textbf{Finetune-Reset:} initialize each new task from the immediately previous task's parameters, but reset optimizer state and rollout state; 
(v) \textbf{EWC}: transfer-style initialization with diagonal-Fisher regularization \citep{kirkpatrick2017overcoming}; and (vi) \textbf{L2Init}: transfer-style initialization with an initialization-preserving penalty \citep{kumar2023maintaining}; (vii) \textbf{DFF}: replaces the standard ReLU with Deep Fourier Features to improve plasticity and sustain learning \citep{lewandowskiplastic}; (viii) \textbf{Shrink-and-Perturb}: transfer-style initialization where, at each task boundary, the previous task's parameters are shrunk and then perturbed with small random noise before training on the next task \citep{ash2020warm}; and (ix) \textbf{TeLAPA-Static:} an ablated version of our method that uses the same shared latent space but keeps the embedder fixed and disables the online re-embedding procedure.

\vspace{-6pt}
\begin{table*}[ht]
    \centering
    \small
    \begin{tabular}{lp{1.60cm}cccccc}
        \toprule
        Method & Mean SR $\uparrow$ & TTT $\downarrow$ & BWT $\uparrow$ & Coverage $\uparrow$ & nBWT $\uparrow$ & TR $\uparrow$ \\
        \midrule
        \textsc{Scratch}
        & $0.374 \pm 0.03$ 
        & $7.75 \pm 2.19$ 
        & $-0.283 \pm 0.052$ 
        & $0.33 \pm 0.04$ 
        & $-0.748 \pm 0.106$ 
        & $0.02 \pm 0.03$ \\

        \textsc{Scratch-Reuse} 
        & $0.525 \pm 0.02$ 
        & $5.49 \pm 2.33$ 
        & $-0.054 \pm 0.008$ 
        & $0.30 \pm 0.03$ 
        & $\mathbf{0.009 \pm 0.001}$ 
        & $0.24 \pm 0.04$ \\

        \textsc{Finetune}
        & $0.588 \pm 0.09$ 
        & $6.55 \pm 2.42$ 
        & $-0.262 \pm 0.065$ 
        & $0.37 \pm 0.04$ 
        & $-0.695 \pm 0.146$ 
        & $0.14 \pm 0.05$ \\

        \textsc{Finetune-Reset} 
        & $0.576 \pm 0.10$ 
        & $5.91 \pm 2.15$ 
        & $-0.263 \pm 0.073$ 
        & $0.33 \pm 0.06$ 
        & $-0.709 \pm 0.147$ 
        & $0.15 \pm 0.04$ \\

        \textsc{EWC}
        & $0.487 \pm 0.06$ 
        & $5.58 \pm 2.53$ 
        & $\mathbf{-0.014 \pm 0.048}$ 
        & $0.41 \pm 0.09$ 
        & $-0.021 \pm 0.055$ 
        & $0.05 \pm 0.03$ \\

        \textsc{L2Init}
        & $0.178 \pm 0.04$ 
        & $9.54 \pm 2.06$ 
        & $-0.134 \pm 0.031$ 
        & $0.11 \pm 0.06$ 
        & $-0.469 \pm 0.197$ 
        & $0.02 \pm 0.03$ \\

        \textsc{DFF} 
        & $0.388 \pm 0.01$  
        & $8.04 \pm 2.26$ 
        & $-0.101 \pm 0.009$  
        & $0.09 \pm 0.04$ 
        & $-0.212 \pm 0.068$ 
        & $0.01 \pm 0.02$  \\

        \textsc{$S\&P$} 
        & $0.586 \pm 0.08$  
        & $6.30 \pm 2.20$  
        & $-0.291 \pm 0.065$ 
        & $0.40 \pm 0.06$ 
        & $-0.414 \pm 0.113$ 
        & $0.12 \pm 0.06$  \\

        \textsc{TeLAPA-Static} 
        & $0.507 \pm 0.10$   
        & $5.59 \pm 2.49$ 
        & $-0.294 \pm 0.059$ 
        & $0.38 \pm 0.09$ 
        & $-0.702 \pm 0.063$ 
        & $0.12 \pm 0.05$ \\

        \textsc{TeLAPA} 
        & $\mathbf{0.706 \pm 0.08}$  
        & $\mathbf{3.35 \pm 2.09}$  
        & $-0.303 \pm 0.086$ 
        & $\mathbf{0.50 \pm 0.09}$ 
        & $-0.629 \pm 0.071$ 
        & $\mathbf{0.25\pm 0.05}$  \\
        \bottomrule
    \end{tabular}
    \vspace{-0.5em}
    \caption{
    \textbf{\textsc{TeLAPA} against single-model baselines.}
    Across 20 runs (mean $\pm$ 95\% CI), \textsc{TeLAPA} is strongest on metrics most directly tied to continual recoverability---higher mean SR, faster TTT, broader task coverage, and stronger threshold retention---supporting the claim that preserving policy neighborhoods improves later reuse relative to single-model preservation. 
    Mean SR, Coverage, nBWT, and TR are higher-is-better; TTT is reported in millions of environment steps and lower-is-better. 
    For BWT, values closer to 0 are better, negative values indicate stronger forgetting.
    }
    \label{tab:results_main_baselines}
    \vspace{-6pt}
\end{table*}

Together, these baselines let us ask not just whether \textsc{TeLAPA} works better, but why: whether its advantage comes from preserving neighborhoods rather than a single model, from retaining more than one solution per task, and from improving recoverability rather than only immediate reuse.
Table~\ref{tab:results_main_baselines} shows that \textsc{TeLAPA} is the strongest overall method on the metrics most directly tied to continual recoverability. It achieves the highest mean success rate ($0.706 \pm 0.08$), the lowest standardized time-to-threshold ($3.35 \pm 2.09$M steps), the highest coverage ($0.50 \pm 0.09$), and the highest threshold retention ($0.26 \pm 0.04$). Relative to \textsc{Finetune} and \textsc{Finetune-Reset}, this indicates that preserving and navigating a neighborhood of competent policies is more effective than carrying forward a single evolving model. 
Crucially, \textsc{TeLAPA} also outperforms \textsc{Scratch-Reuse}, which stores one solution per task, improving mean SR from $0.525$ to $0.706$ and reducing TTT from $5.49$M to $3.35$M steps. 
This suggests that the gain cannot be explained by task-indexed caching alone, but depends on retaining and selecting among multiple nearby competent alternatives. 
At the same time, \textsc{Scratch-Reuse} remains best on BWT and nBWT, so \textsc{TeLAPA}'s advantage is not best interpreted as strict no-forgetting of a single stored policy. 
Rather, its advantage is that it learns more tasks in the first place and recovers them more effectively when they reappear. 
This interpretation is further supported by the gap between \textsc{TeLAPA-Static} and full \textsc{TeLAPA}, which shows that archive storage alone is insufficient: online maintenance of the shared latent geometry materially improves continual transfer.
However, \textsc{TeLAPA}'s gains come with measurable overhead, primarily due to archive illumination, few-shot origin selection, and latent-space maintenance. 
\textsc{TeLAPA} required $1.30 \pm 0.10\times$ the wall-clock time of \textsc{Finetune} and reached $53.33 \pm 1.54$GB peak CPU RAM. 
We quantify the full associated compute-memory tradeoff in App.~\ref{app:compute_memory_overhead}.
\vspace{-0.5em}
\subsection{Local transfer reveals failure mode of single-model preservation}
\label{sec:true_local_envelope}
\vspace{-0.5em}
The baseline comparison above showed that preserving policy neighborhoods improves continual recoverability relative to single-model baselines, including methods that reuse or store only one policy per task. 
We now directly test the mechanism behind that advantage. 
In the most favorable version of single-model preservation, one would retain the source-optimal policy and reuse it later as the transfer seed. 
The question in this subsection is whether that single representative is actually sufficient, or whether nearby source-competent alternatives---policies that still perform well on the source task---already contain transfer-relevant variation that a one-policy summary discards.

We show a consistent local failure mode of single-model preservation separated across three questions: 
(i) Does the source-competent set contain multiple viable alternatives, or does it collapse to a single latent point? 
(ii) Even near the preserved source-best elite, do competent alternatives differ in their downstream transfer utility? 
(iii) Does searching for this local competent basin recover better transfer policies than preserving only the source-best elite?

Let $\mathcal{C}_{s \rightarrow t}$ denote the set of policies from source task $s$ evaluated as transfer for target task $t$.
For each candidate policy $\pi_i \in \mathcal{C}_{s \rightarrow t}$, we record its source-task fitness $f_i^{src}$, its target performance after few-shot transfer evaluation $y_i^{tgt}$, and its latent embedding $z_i$ in the shared behavior space. 
We define the \emph{source-best} elite in $\mathcal{C}_{s \rightarrow t}$ as
$
\pi^{*}_{src} = \arg\max_{\pi_i \in \mathcal{C}_{s \rightarrow t}} f_i^{src}
$
and use its latent embedding $z^{*}_{src}$ as the reference point for the local basin analysis. 
This design directly matches the single-model preservation setting, since such methods would naturally retain the source-optimal representative solution.


\vspace{-8pt}
\paragraph{Competent source set.}
We first define the \emph{good-enough} source set $\mathcal{G}_{s \rightarrow t}$ as the set of elite policies ($\pi_i$) for which its source task fitness $f_i^{src} \ge \gamma \cdot f_{src}^{*}$ where $f_{src}^{*}$ is the source-best elite fitness and $\gamma \in (0,1]$ is the source-competence threshold. This set isolates the candidates that would still be considered acceptable source solutions, while allowing multiple alternatives to coexist.

\vspace{-8pt}
\paragraph{Local basin inside the competent set.}
We define locality \emph{within} the competence set. 
For every $\pi_i \in \mathcal{G}_{s \rightarrow t}$, we compute its latent distance to the source-best elite: $d_i = \| z_i - z^{*}_{src} \|_2$,
and rank all candidates by increasing distance to $z^{*}_{src}$. Let $\rho_i \in [0,1]$ denote this normalized distance-rank, where $\rho_i = 0$ corresponds to the closest competent candidate and $\rho_i = 1$ to the farthest competent candidate. Then, we define the \emph{competent local basin} $\mathcal{B}_{s \rightarrow t}$ as the set of elites for which $\rho_i \le \tau$ where $\tau$ is the local-rank threshold. 


\begin{wrapfigure}{r}{0.32\textwidth}
    \vspace{-2em}
    \centering
    \includegraphics[width=\linewidth]{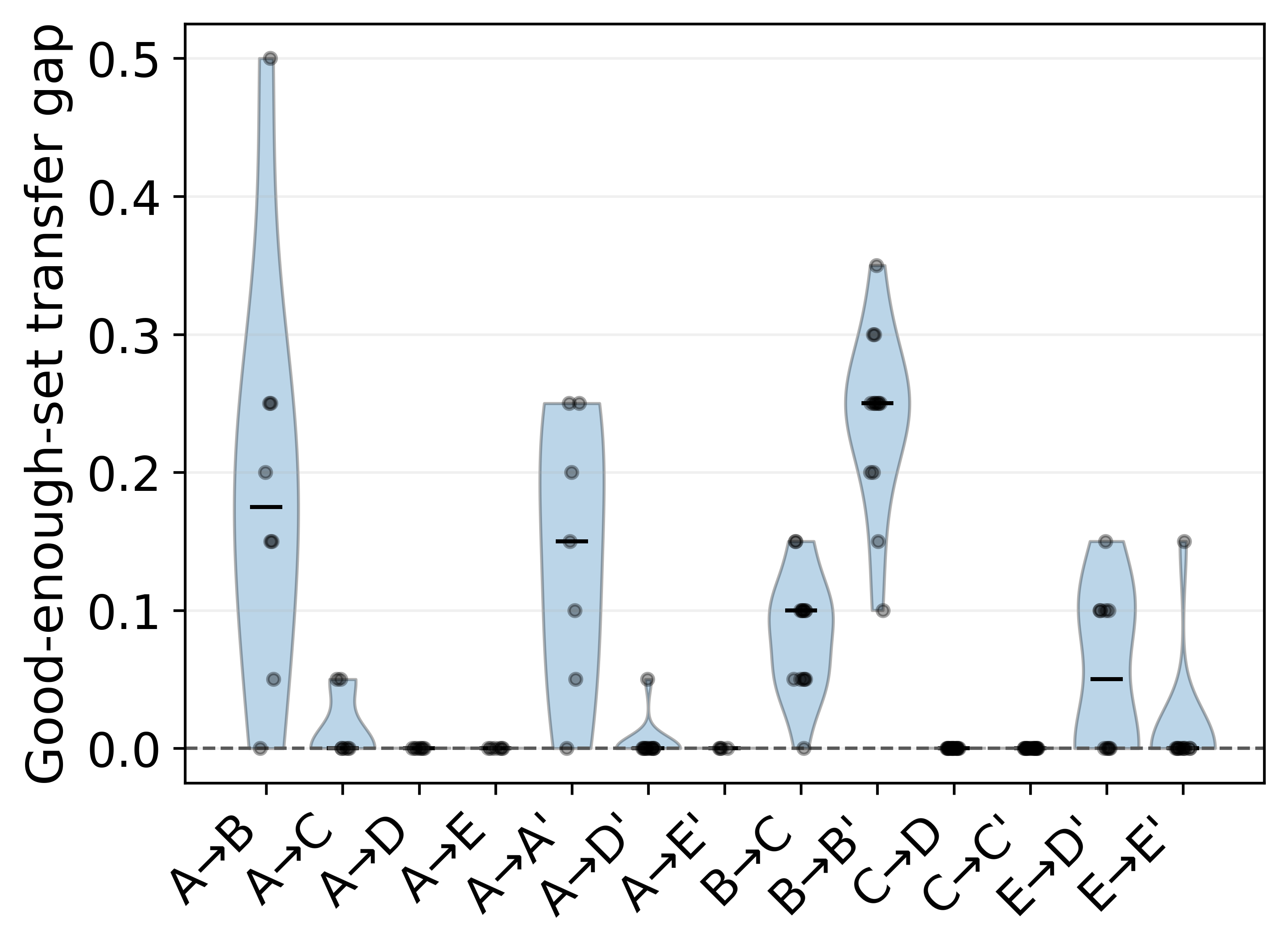}
    \vspace{-2em}
    \caption{Transfer gap in target task performance between the best \emph{good-enough} elite and source-optimal elite.}
    \label{fig:local_transfer_panel_a}
    \vspace{-16pt}
\end{wrapfigure}

\vspace{-8pt}
\paragraph{Source-best is not always transfer-best.}
We demonstrate the failure of single-model preservation and show that source-optimality does not uniquely determine transfer-optimality.  
For each $s \rightarrow t$ pair, we compare the target performance of the source-best elite against the policy with the highest target performance within the entire \emph{good-enough} set:
\[
\Delta_{\mathrm{good}}
=
\max_{\pi_i \in \mathcal{G}_{s \rightarrow t}} y_i^{tgt}
-
y^{tgt}(\pi^{*}_{src}).
\]
Fig.~\ref{fig:local_transfer_panel_a} shows that the source-best elite is often not the transfer-best elite even among source-competent alternatives: in $35.5\%$ of all $s \rightarrow t$ pairs, the best policy in the \emph{good-enough} set outperforms the preserved source-best elite on the target (transfer gap $> 0$). 
Thus, even before imposing locality, a method that preserves only $\pi^{*}_{src}$ already discards better downstream options that were available in the source archive. 
This directly tests the intuition behind the illumination phase: if local perturbations only produced redundant copies or unstable failures, then the source-competent set would collapse near the source-best elite and provide little downstream transfer variation.

\begin{wrapfigure}{r}{0.32\textwidth}
    \vspace{-22pt}
    \centering
    \includegraphics[width=\linewidth]{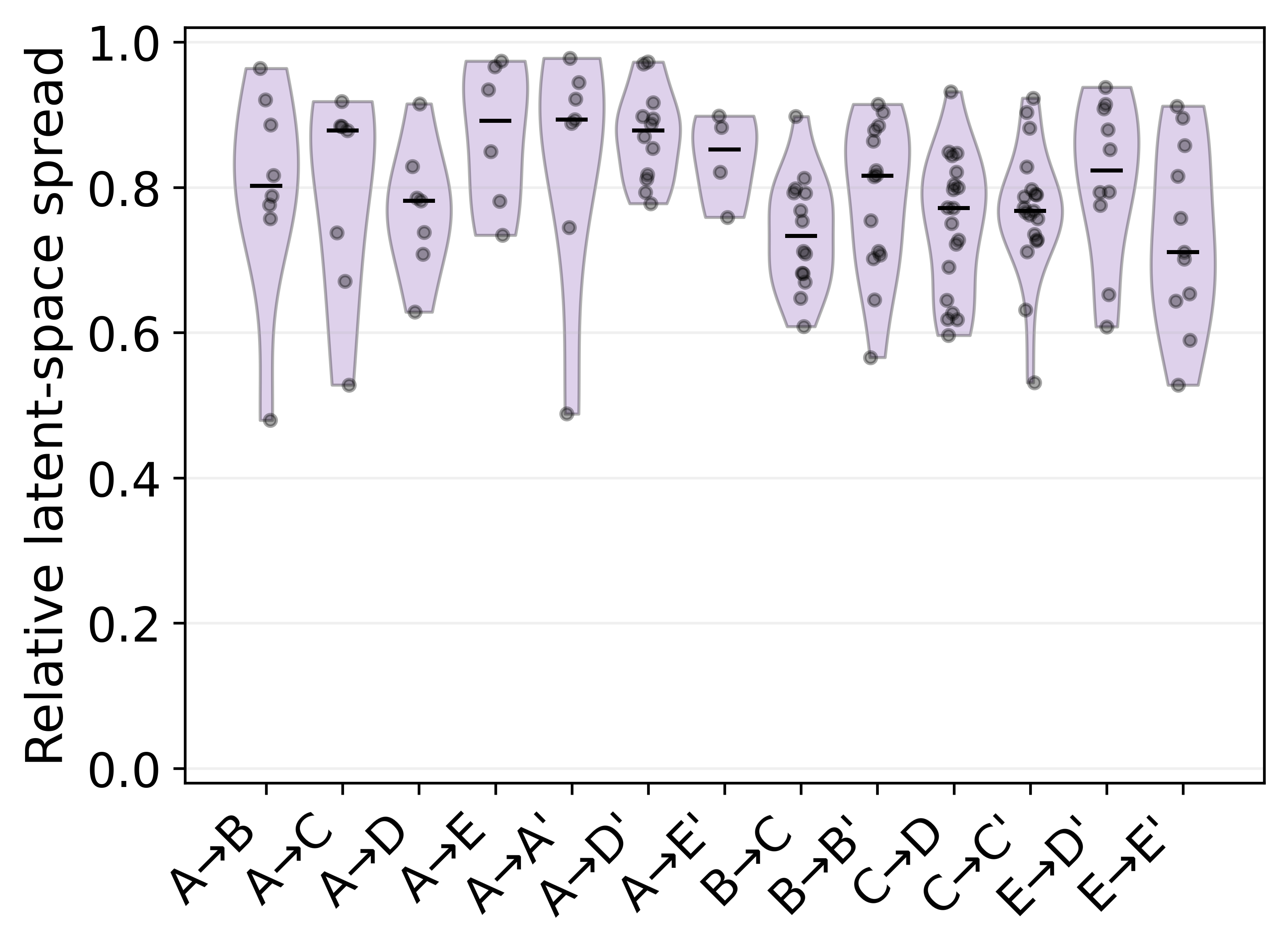}
    \vspace{-2em}
    \caption{Relative latent spread of the \emph{good-enough} set.}
    \label{fig:local_transfer_panel_b}
    \vspace{-10pt}
\end{wrapfigure}

\vspace{-5pt}
\paragraph{The good-enough set occupies a broad latent basin.}
If the \emph{good-enough} set were concentrated at essentially one point, then preserving a single representative policy could indeed be sufficient. Fig.~\ref{fig:local_transfer_panel_b} tests this by measuring how broadly the \emph{good-enough} set spreads around the source-best elite relative to the full candidate set: 
\[
\mathrm{Span}_{0.75}
=
\frac{
Q_{0.75}\left( \left\{ \|z_i - z^{*}_{src}\|_2 : \pi_i \in \mathcal{G}_{s \rightarrow t} \right\} \right)
}{
\max_{\pi_j \in \mathcal{C}_{s \rightarrow t}} \|z_j - z^{*}_{src}\|_2
},
\]
where $Q_{0.75}(\cdot)$ denotes the $75$th percentile. A high value means that most source-competent candidates extend far from $\pi^{*}_{src}$ relative to the full available candidate range, rather than collapsing into a narrow neighborhood. 
In our analysis, the median $\mathrm{Span}_{0.75}$ is $0.792$, showing that the \emph{good-enough} set occupies a substantial portion of the candidate latent region. 
This indicates that the archive is not merely storing redundant source-competent copies of the same policy, but retains a broadly distributed set of behaviorally distinct candidates. As a result, transfer from the archive is likely to expose multiple qualitatively different source-competent behaviors, any of which may later outperform the source-optimal policy on the target task.

\begin{wrapfigure}{r}{0.32\textwidth}
    \vspace{-5pt}
    \small
    \includegraphics[width=\linewidth]{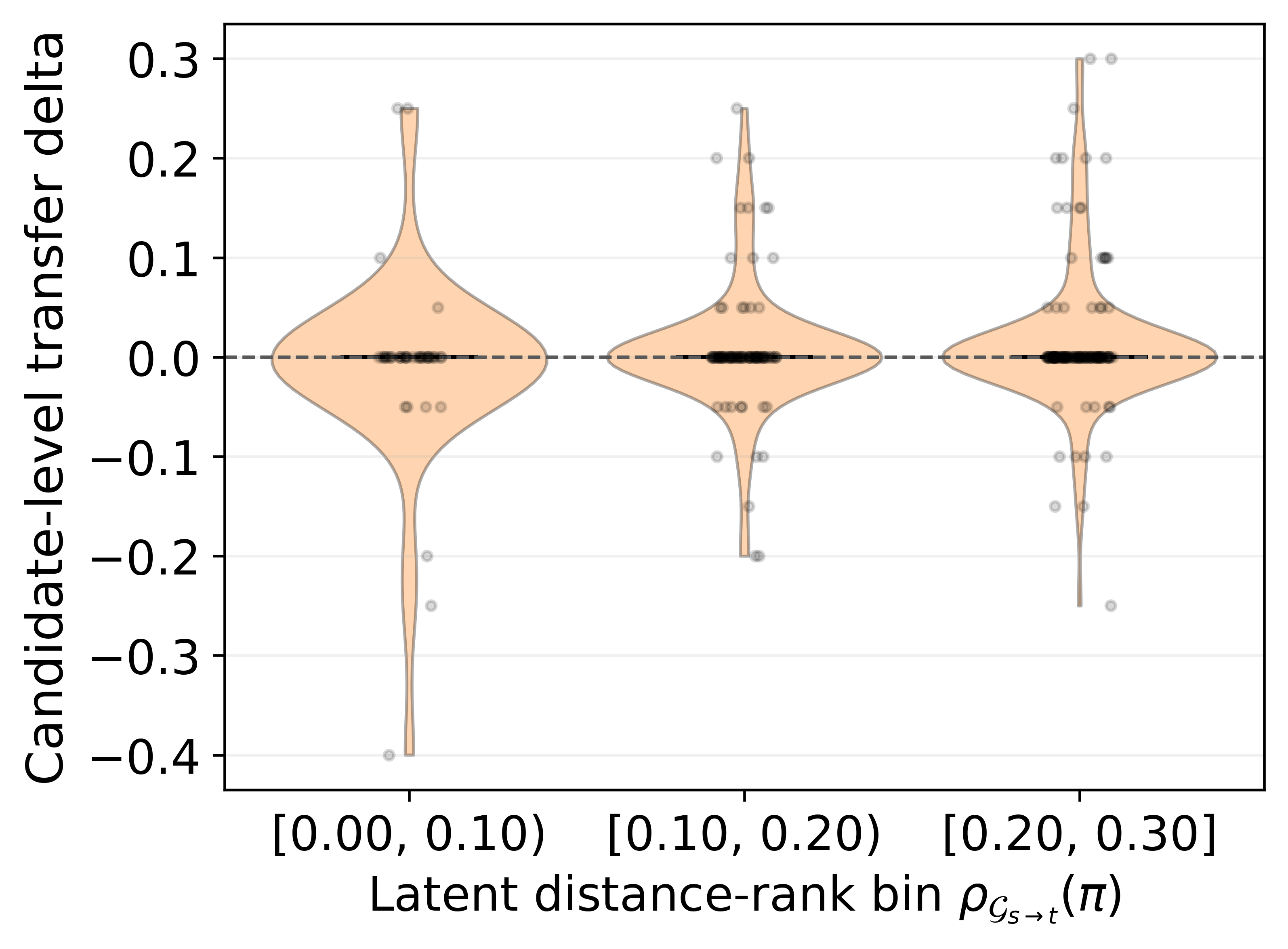}
    \vspace{-2em}
    \caption{Latent distance-rank bins, relative to the source-best, pooled within the local competent basin $\mathcal{B}$.}
    \label{fig:local_transfer_panel_c}
    \vspace{2pt}
\end{wrapfigure}

\vspace{-0.5em}
\paragraph{Transfer differs even within a local competent neighborhood.}
A stricter test is whether the source-best policy is already insufficient even within its own local competent neighborhood, rather than only because better transfer seeds exist farther away in the source-competent set.
To test this, Fig.~\ref{fig:local_transfer_panel_c} restricts attention to the competent local basin $\mathcal{B}_{s \rightarrow t}$ around the source-best elite and compares each candidate $\pi_i \in \mathcal{B}_{s \rightarrow t}$ against that preserved source-best policy through
\[
\delta_i = y_i^{tgt} - y^{tgt}(\pi^{*}_{src}).
\]
Candidates are grouped by latent distance-rank $\rho_i$ within the \emph{good-enough} set, so the panel asks whether nearby source-competent alternatives are effectively interchangeable for downstream transfer. 
They are not. 
If these nearby source-competent candidates were effectively interchangeable, then $\delta_i$ would concentrate near zero and the source-best elite would be a good proxy for its neighborhood. Instead, across rank bins, the distributions of $\delta_i$ remain centered near zero but retain clear positive and negative tails, indicating that policies that are all source-competent and locally nearby can still differ meaningfully in target performance.
Thus, even within a tight competent neighborhood around the retained source-best elite, nearby policies are not interchangeable for downstream transfer. 
The point is not that the transfer-optimal seed must also be local---it may still lie farther away elsewhere in the archive---but that meaningful transfer variation already exists in the immediate neighborhood of the policy a single-model method would retain.

\begin{wrapfigure}{r}{0.32\textwidth}
    \vspace{-20pt}
    \centering
    \includegraphics[width=\linewidth]{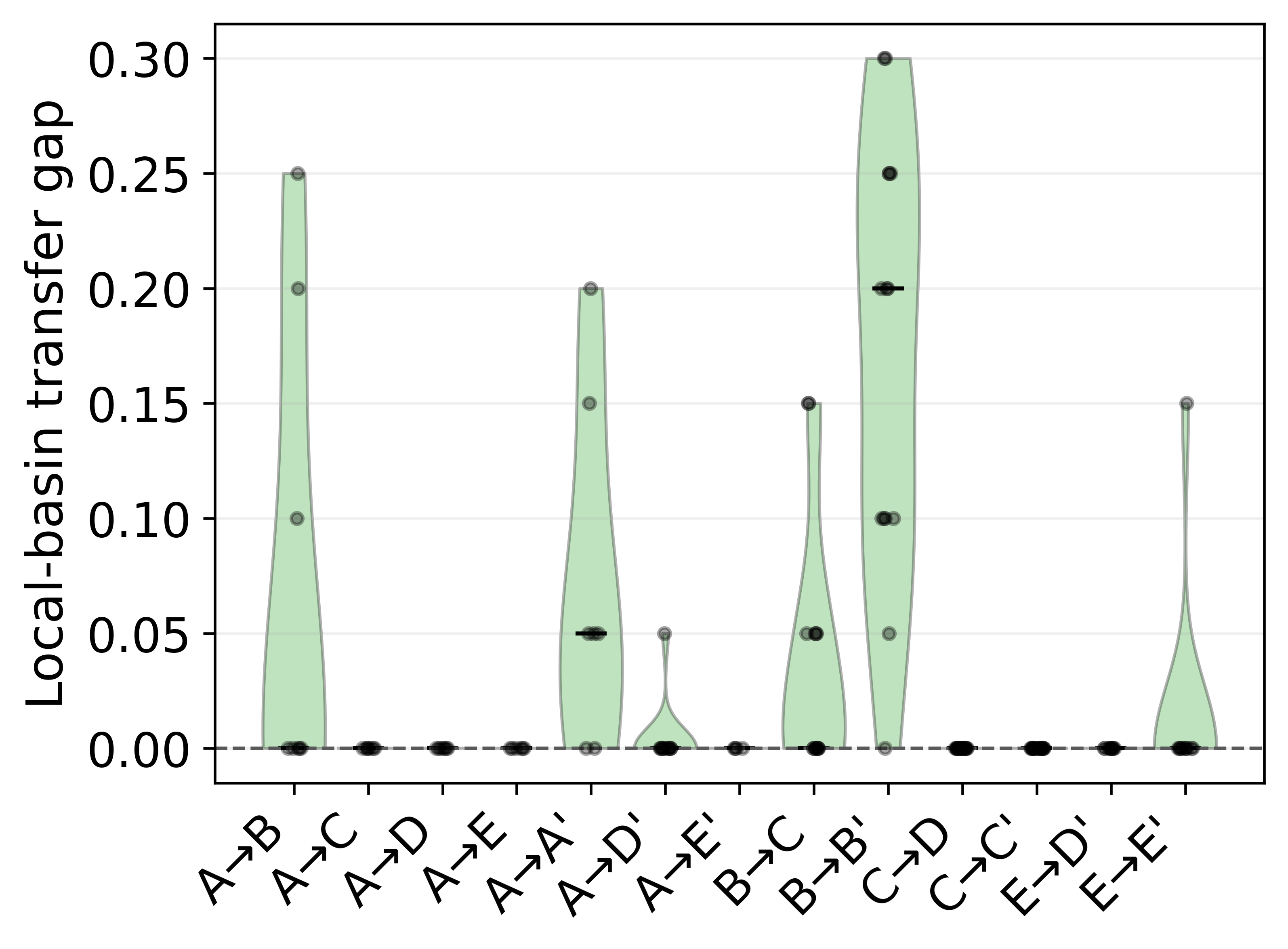}
    \vspace{-2em}
    \caption{Transfer gain obtained by searching only within the local basin.}
    \label{fig:local_transfer_panel_d}
    \vspace{-15pt}
\end{wrapfigure}

\vspace{-5pt}
\paragraph{Collapsing the local basin to one preserved policy incurs a transfer cost.}
Fig.~\ref{fig:local_transfer_panel_d} summarizes the practical consequence of the local heterogeneity shown in Fig.~\ref{fig:local_transfer_panel_c}. 
We compare the preserved source-best elite against the best candidate available inside the competent local basin:
$
\Delta_{\mathrm{local}}
=
\max_{\pi_i \in \mathcal{B}_{r,s \rightarrow t}} y_i^{tgt}
-
y^{tgt}(\pi^{*}_{src}).
$
This occurs in a substantial fraction of $s \rightarrow t$ pairs, showing that the competent neighborhood around the source-best elite is not transfer-redundant. 
Archive-based preservation is therefore valuable not just for maintaining global diversity, but for retaining nearby competent alternatives with distinct downstream utility. 
Although some transfer-optimal policies may lie outside this tight local basin, the local result alone shows that reducing the preserved neighborhood to a single representative is already suboptimal.
\vspace{-0.5em}
\subsection{Stepping-Stone Lineages Across Task Archives}
\label{subsec:stepping_stones}
\vspace{-0.5em}
Previously, we showed that single-model preservation methods retain only one policy snapshot, thereby collapsing alternative behaviors that may later prove more adaptable under new tasks or revisits. 
\textsc{TeLAPA} addresses this limitation by preserving an archive of competent, behaviorally distinct policies, including intermediate ones that may not be source-optimal but still retain useful structure from which later high-performing solutions can emerge.
Now, we ask whether this effect extends beyond a single local neighborhood to the archive system as a whole. 
Specifically, we study whether \textsc{TeLAPA} preserves a longer-horizon \emph{stepping-stones} structure: intermediate policies that are not necessarily final or source-optimal, but whose archived task history makes them especially reusable later. 
This view is closely related to novelty search and quality-diversity, which argue that search benefits from preserving diverse intermediates rather than collapsing onto a single objective-maximizing solution \citep{lehman2011abandoning, mouret2015illuminating, gaier2019stepping, nordmoen2021map}.
\textsc{TeLAPA} repeatedly selects elites from prior archives, adapts them to new tasks, and stores them again as future transfer candidates. 
Accordingly, what matters is not only which archive supplies a policy to the current target, but also how that policy was shaped by earlier transfers. 
Therefore, our goal is not to reconstruct the full optimization trace of every elite, but to recover the minimal lineage structure needed to study how reusable policies move through task space over time.

\begin{figure*}[ht]
\vspace{-0.5em}
  \centering
  \includegraphics[width=\textwidth]{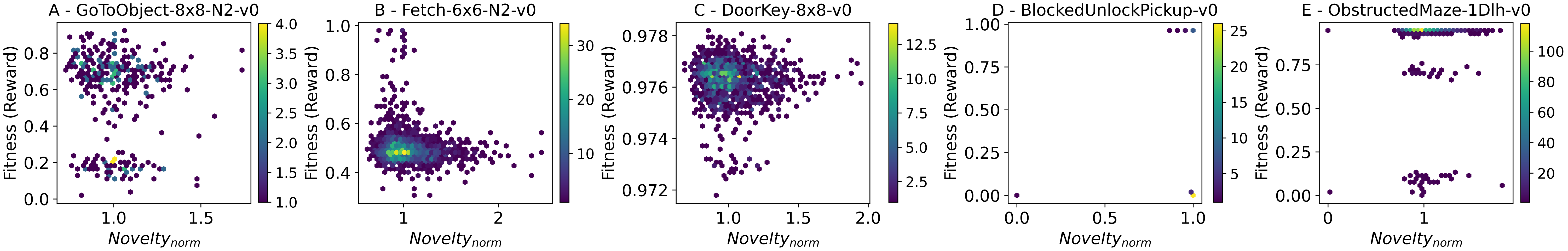}
  \vspace{-2em}
  \caption{\textbf{Pooled QD structure across runs.}
    We visualize the density of archive elites as a function of fitness and normalized novelty \(\nu^{\mathrm{norm}}\).
    Small novelty indicates local redundancy, whereas large novelty indicates more isolated or frontier elites.
    Differences across tasks reveal how each archive balances competence and behavioral spread within its own latent geometry.
    }
  \label{fig:pooled_qd_hexbin}
  \vspace{-15pt}
\end{figure*}

\vspace{-8pt}
\paragraph{Intra-task archive structure.}
Before analyzing cross-task lineages, we first characterize the internal geometry of each task archive. For elite \(i\) in task \(T_k\), we define a novelty proxy as the within-task nearest-neighbor distance in standardized latent space,
$\nu^{\mathrm{norm}}_{k,i} \;=\; \frac{\min_{j \neq i}\lVert \tilde{z}_{k,i} - \tilde{z}_{k,j}\rVert_2}{\mathrm{median}_i(\nu_{k,i})+\epsilon}.$
Figure~\ref{fig:pooled_qd_hexbin} shows that these archives do not collapse to single redundant clusters, but instead exhibit task-dependent internal structure. 
Tasks A and B show broad support across novelty with only modest fitness degradation, indicating that competent but behaviorally distinct elites coexist within each archive. 
Task C shows the clearest case of relatively ``cheap'' diversity, maintaining a concentrated high-fitness region across a wider novelty range. 
Task E is more stratified, multiple fitness bands coexist across novelty, suggesting a rougher internal geometry with several behavioral regimes rather than one dominant mode. 
Task D is too sparse to support a strong within-task interpretation. 
Overall, the archive library contains both dense local groupings and more isolated frontier elites, providing the internal heterogeneity needed for later retrieval and stepping-stone reuse.

\begin{figure*}[ht]
\vspace{-0.5em}
    \centering
    \includegraphics[width=\textwidth]{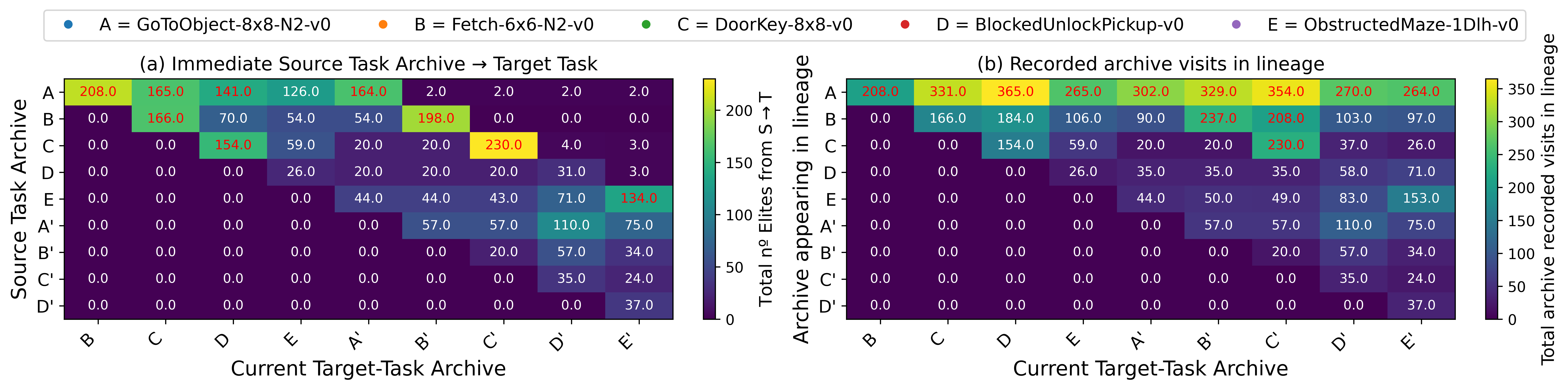}
    \vspace{-2em}
    \caption{
    \textbf{Stepping-stone lineage structure across task archives.}
    \textbf{(a)} \textit{Immediate source-task archive \(\rightarrow\) current target-task archive.}
    Rows denote the most recent source archive used for transfer, and columns denote the current target-task archive. 
    Each cell shows how many evaluated candidates for that target were drawn from that immediate source archive.
    \textbf{(b)} \textit{Recorded archive visits in lineage.}
    Columns denote the current target-task archive, and rows denote archive tags appearing anywhere in recorded lineage history. 
    Each cell shows the total number of recorded lineage visits to archive tag \(S\) among candidates evaluated for target archive \(T\). 
    Counts are overlapping rather than exclusive: a single candidate contributes to every archive tag that appears in its lineage.
    }
    \label{fig:stepping_stones}
    \vspace{-10pt}
\end{figure*}

\vspace{-0.8em}
\paragraph{Cross-task lineage structure.}
Figure~\ref{fig:stepping_stones} shows that these non-redundant within-task archives are not used only as isolated task memories, but participate in longer multi-archive lineages. 
Panel~(a) summarizes \emph{immediate} transfer reuse: for each current target-task archive, it shows how many evaluated candidates were drawn from each most-recent source archive. 
Reuse is not strictly sequential: rather than drawing only from the immediately preceding task, several later targets receive candidates from a broader historical library, including both original-task and revisit archives. 
Recall that when a new target task arrives, \textsc{TeLAPA} constructs its transfer candidate pool from the union of all previously learned task archives, so any earlier archive can in principle supply a transfer seed. Thus, later reuse is not constrained to follow the curriculum order, but can be shaped by downstream utility across the full archive library.
While panel~(a) focuses on immediate source selection, panel~(b) summarizes deeper recorded ancestry. 
For candidates evaluated for the current target-task archive on each column, each cell counts how often the archive tag on the row appears anywhere in their recorded lineage history. 
Thus, a large value in row \(D'\), column \(E\) does not indicate backward-in-time transfer from \(E\) to \(D'\); it indicates that candidates evaluated for \(E\) often carry recorded ancestry that passes through \(D'\). 
These lineage counts show that later target archives are not shaped only by their most recent transfer source. Instead, candidates evaluated on later targets often retain ancestry through multiple earlier archive tags, so reuse reflects accumulated historical paths rather than only fresh one-hop transfers. 
Early archive tags continue to appear across many later targets, while intermediate tags recur in patterns consistent with historical bridge roles. 
Together, these patterns indicate that \textsc{TeLAPA} maintains multi-archive stepping-stone lineages rather than collapsing reuse to a single representative per task.

The presence of multi-stage ancestry suggests that transfer on a target task can depend on more than its immediate source archive, and may instead reflect a longer historical path through earlier tasks (e.g., success on \(C\) may depend not only on the immediate source \(B\), but on the broader trajectory \(A \rightarrow B \rightarrow C\)). I
n this sense, \textsc{TeLAPA} supports an internal \emph{skill-aligned curriculum}, where later reuse is shaped not only by the most recent transfer step but also by deeper lineage structure preserved across archives. However, structural depth alone is not yet a functional explanation. 
Therefore, we next analyze which lineage properties are actually associated with improved later reuse.

\begin{figure*}[ht]
    \centering
    \includegraphics[width=\textwidth]{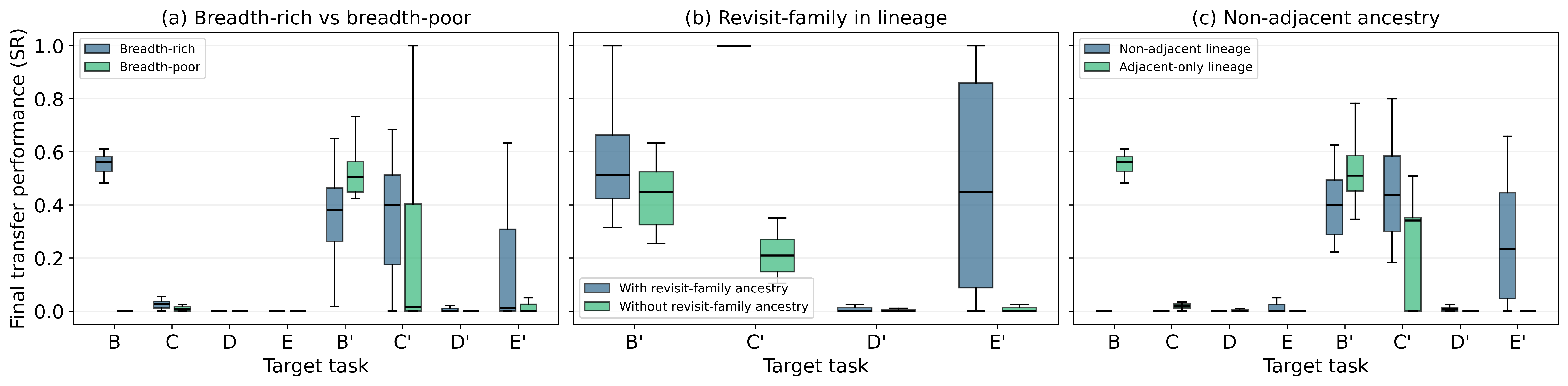}
    \caption{
    \textbf{Lineage utility across transferred candidates.}
    Each panel compares the distributions of final transfer performance for two candidate groups defined by a different lineage property, shown separately for each target task.
    \textbf{(a)} \textit{Breadth-rich versus breadth-poor lineage histories}. 
    Candidates are grouped by whether their lineage contains the original archive family corresponding to the current target.
    \textbf{(b)} \textit{Revisit-family membership}. 
    Candidates are grouped by whether their lineage contains the original family corresponding to the current target. We measure whether including that target-relevant family predicts higher final transfer performance.
    \textbf{(c)} \textit{Adjacent versus non-adjacent ancestry}. 
    Candidates are grouped by whether their lineage includes a non-adjacent transition in archive-family order.
    }
    \label{fig:lineage_utility}
    \vspace{-15pt}
\end{figure*}

\vspace{-0.5em}
\paragraph{Which lineage properties predict later reuse?}
Structural existence alone does not imply functional importance. 
Fig.~\ref{fig:lineage_utility} asks whether simple lineage properties are associated with higher final transfer performance by comparing, for each target, the distributions of final transfer success across two lineage-defined candidate groups. 
Panel~(a) tests generic \emph{breadth}, defined as the number of distinct archive families in a candidate's history. 
Breadth-richer lineages help for some targets, especially several revisits, but the effect is weak or inconsistent for others, showing that generic historical breadth is not by itself a universal explanation for reuse.
Panel~(b) tests \emph{task-relevant} ancestry: candidates are divided by whether their lineage contains that target's original archive family. 
For several revisits, candidates whose history includes the corresponding original family tend to perform better, suggesting that reuse depends more on preserving specific task-relevant stepping stones than on accumulating broad lineage history alone. 
Thus, \textsc{TeLAPA}'s advantage lies not just in maintaining long-lived lineages, but in preserving lineage structure that remains useful for future reuse.
Panel~(c) tests whether useful stepping stones must follow only adjacent curriculum progression by comparing candidates with and without a non-adjacent transition in their lineage. 
Although the effect varies across targets, non-adjacent ancestry is beneficial for several revisits, indicating that useful stepping stones need not lie only along the shortest curriculum path. 
This supports the stepping-stone view: \textsc{TeLAPA}'s efficacy does not come from any single lineage pattern being uniformly best, but from maintaining a diverse \emph{behavioral library} of reusable histories.

\vspace{-5pt}

\section{Conclusion \& Future Work}
\vspace{-0.5em}
\paragraph{Conclusion}
Throughout this paper, we argued that continual reinforcement learning should look beyond approaches that rely on a single evolving policy as the main reusable solution across tasks. 
Under non-stationarity and interference, keeping one previously successful policy available is often not enough to support future adaptation. 
Instead, later reuse can depend on access to a structured set of competent, behaviorally related alternatives from which rapid re-adaptation can emerge. 
Motivated by this view, we introduced \textsc{TeLAPA}, a continual RL framework that organizes behaviorally diverse policy neighborhoods into per-task archives and maintains a shared latent space so that archived policies remain comparable and reusable under non-stationary drift.
Across our MiniGrid setting, this perspective led to a consistent empirical picture. 
\textsc{TeLAPA} achieved the strongest overall performance on the metrics most directly tied to learning, re-adaptation, and retention.
Our analyses further showed why relying on a single policy is insufficient: even source-competent policies near the source optimum can differ meaningfully in transfer value, and effective reuse depends not only on retaining task-specific candidates, but also on maintaining stepping-stone structure across archives. 
Taken together, these results suggest that sustaining functional plasticity in continual RL depends on maintaining a structured, skill-aligned archive of diverse policies rather than collapsing reuse to a single successful policy.
\vspace{-5pt}
\paragraph{Future Work}
Future work should test whether these ideas hold beyond the cycle-based MiniGrid setting, including longer task streams, weaker task boundaries, richer visual domains, continuous-control environments, and settings where the interaction rules themselves change. 
It should also identify which forms of diversity are genuinely useful by isolating the causal roles of archive construction, latent-space maintenance, and lineage formation in later transfer. Because retrieval depends on distances in a shared behavior space, latent-space maintenance should be treated as a central problem in its own right, with methods that directly preserve navigability and cross-task discriminability under non-stationary drift. 
More broadly, a natural next step is to move beyond archive-augmented PPO toward continual RL algorithms that jointly optimize representation learning, archive growth, and transfer-time selection around future adaptability rather than only immediate performance or parameter retention.

\newpage
\section*{Acknowledgments}
This material is based on work supported by the National Science Foundation under Grant No. 2218063 and 2239691. The authors acknowledge the Vermont Advanced Computing Center (VACC) at the University of Vermont for providing computational resources that have contributed to the research results reported in this paper.

\bibliography{collas2026_conference}
\bibliographystyle{collas2026_conference}

\newpage
\appendix

    

\section{MiniGrid: Environment Suite and Curriculum Design}
\label{app:envs}

\subsection{MiniGrid tasks used in the curriculum}
\label{app:env_list}

We evaluate continual transfer on a MiniGrid curriculum designed as a \emph{skill ladder}: each task introduces a small number of new interaction requirements while reusing earlier ones.
This makes transfer \emph{falsifiable}: if a method preserves adaptable skill neighborhoods, revisits should recover faster and with higher success than training from scratch.

\begin{table}[h]
\centering
\small
\setlength{\tabcolsep}{6pt}
\begin{tabular}{clp{0.5\linewidth}}
\toprule
\textbf{Tag} & \textbf{Environment ID} & \textbf{Primary skills and incremental difficulty} \\
\midrule
A & \texttt{MiniGrid-GoToObject-8x8-N2-v0} &
Navigation and text-conditioned goal specification (grounding color/type); no object pickup required. Establishes basic exploration, obstacle avoidance, and mission conditioning. \\

B & \texttt{MiniGrid-Fetch-6x6-N2-v0} &
Adds object interaction: pick up the correct object specified by the mission and bring it to a target condition (delivery).
Requires learning the \texttt{pickup} primitive and linking it to text grounding, beyond pure navigation. \\

C & \texttt{MiniGrid-DoorKey-8x8-v0} &
Introduces the key--door dependency: obtain a key, unlock/open a door, then navigate to the goal.
Requires sequential subgoals (\texttt{pickup key} $\rightarrow$ \texttt{toggle/open door} $\rightarrow$ \texttt{reach goal}) under cluttered navigation. \\

D & \texttt{MiniGrid-BlockedUnlockPickup-v0} &
Adds clutter interaction and ``don’t pick up the wrong thing'' pressure.
The agent must manipulate blocking objects to access the relevant interaction chain while avoiding detrimental pickups (e.g., moving blockers out of the way without committing to incorrect object handling). \\

E & \texttt{MiniGrid-ObstructedMaze-1Dlh-v0} &
Longest-horizon task: maze-like navigation with obstructing objects and a hidden key, requiring both exploration and correct subgoal chaining.
This setting stresses interference and plasticity most strongly: success requires committing to a precise multi-step plan while handling clutter. \\
\bottomrule
\end{tabular}
\caption{\textbf{Curriculum environments and intended skill increments.}
Each task builds on prior primitives while adding a new interaction requirement (navigation $\rightarrow$ pickup $\rightarrow$ unlocking $\rightarrow$ clutter manipulation $\rightarrow$ long-horizon subgoal chaining).}
\label{tab:envs}
\end{table}

\subsection{Revisits and continual-learning signal}
\label{app:env_revisits}

The evaluation sequence includes revisits:
\[
A\rightarrow B\rightarrow C\rightarrow D\rightarrow E\rightarrow A'\rightarrow B'\rightarrow C'\rightarrow D'\rightarrow E'.
\]
Revisits are identical environment families to their first encounters (same environment IDs, new procedural instances/seeds as applicable).
Because the policy archives for $A,B,C,D,E$ have already been populated in the first pass, revisit performance measures \emph{recovery under interference}: the method must re-enter a suitable behavioral basin and (for few-shot selection) regain competence rapidly.
This makes revisits a direct operational proxy for plasticity preservation in continual RL.

\paragraph{Why this curriculum?}
Purely deeper navigation benchmarks can over-emphasize exploration depth while under-testing interference between \emph{interaction primitives}.
Here, later tasks deliberately introduce conflicts (e.g., when to pick up, when to toggle/open, how to move blockers, when to commit to a long chain), so that sequential training can induce both forgetting and loss of plasticity.
A successful continual learner should therefore show both strong forward transfer (reuse of earlier skills) and rapid recovery on revisits despite intervening tasks.

\subsection{Rewards in MiniGrid}
\label{sec:metrics_primary}

MiniGrid rewards are sparse and given only upon successful termination.
For an episode that succeeds at step count $\texttt{step\_count}$, the return is
\[
R \;=\; 1 - 0.9\cdot \left(\frac{\texttt{step\_count}}{\texttt{max\_steps}}\right),
\]
and failure yields $R=0$.
Thus, mean reward near zero typically indicates \emph{no successful episodes}, rather than slow-but-steady progress.
Accordingly, we report both reward and \textbf{success rate (SR)}, where
\[
\mathrm{SR} \;=\; \frac{1}{N}\sum_{i=1}^N \mathbb{I}\!\left[R_i > 0\right],
\]
i.e., the fraction of evaluation episodes that terminate successfully (positive terminal reward).
SR provides the cleanest ``is it solving'' signal in sparse-reward regimes, while reward captures solution efficiency once success is achieved.

\section{Hyper-Parameters sweeps for Baseline Methods}
\label{app:hyperparam_sweep}

To ensure that each method was evaluated under a reasonable configuration, we conducted a targeted hyperparameter search for methods requiring method-specific tuning. 
Tables~\ref{tab:hp_search_baselines} summarize the explored grids and the final selected configuration used in the main experiments. 
Methods without method-specific hyperparameters---namely \textsc{Finetune}, \textsc{DFF}, \textsc{Finetune-Reset}, \textsc{Scratch}, and \textsc{Scratch-Reuse}---were run without an additional method-level tuning grid.

\begin{table*}[ht]
\centering
\small
\setlength{\tabcolsep}{4pt}
\begin{tabularx}{\textwidth}{@{}
>{\raggedright\arraybackslash}p{2.6cm}
>{\raggedright\arraybackslash}p{2.8cm}
>{\raggedright\arraybackslash}X
>{\raggedright\arraybackslash}p{2.8cm}
@{}}
\toprule
\textbf{Method} & \textbf{Hyperparameter(s)} & \textbf{Explored values} & \textbf{Optimal} \\
\midrule

\textsc{EWC} 
& $\lambda$ 
& $\{10,\;25,\;50,\;100,\;200,\;500,\;1000\}$ 
& $\lambda = 10$. \\

\textsc{EWC} 
& \texttt{Decay} 
& $\{0.8,\;0.9,\;0.95,\;0.99\}$ 
& \texttt{Decay = 0.8}. \\

\textsc{L2Init} 
& $\lambda$ 
& $\{0.001,\;0.005,\;0.01,\;0.05,\;0.1,\;0.25,\;0.5,\;1.0\}$ 
& $\lambda = 0.001$. \\

\textsc{Shrink\&Perturb} 
& $\alpha$ 
& $\{0.7,\;0.8,\;0.9,\;0.95,\;0.98,\;0.99\}$ 
& $\alpha = 0.99$. \\

\textsc{Shrink\&Perturb} 
& \texttt{Noise Scale} 
& $\{0.0001,\;0.0005,\;0.001,\;0.005,\;0.01\}$ 
& \texttt{Scale = 1e-3}. \\

\bottomrule
\end{tabularx}
\caption{Hyperparameter grids explored for baseline methods. For EWC, the main grid was run without critic regularization. Methods with more than one hyperparameter were run for all combinations possible.}
\label{tab:hp_search_baselines}
\end{table*}
\vspace{-1.5em}            
\section{MAP-Elites}
\label{app:map_elites}

\paragraph{MAP-Elites Hyperparameters.} 
We also report the main hyperparameters used for the MAP-Elites illumination stage in able~\ref{tab:map_elites_hparams}. 
These control the size of the archive, the strength and duration of the search process, and the way candidate policies are evaluated before being inserted into the archive. In particular, the archive is initialized with a target size and capacity, candidates are evaluated over multiple episodes, and accepted insertions trigger adaptation of the diversity threshold so that the container remains close to its intended operating size.

\begin{table*}[ht]
\centering
\small
\begin{tabular}{p{4.2cm}p{2.8cm}p{8.2cm}}
\toprule
\textbf{Parameter} & \textbf{Value used} & \textbf{Role in the algorithm} \\
\midrule
Archive target size 
& $256$ 
& Desired number of elites to preserve in the archive during illumination. \\

Archive capacity 
& $384$ 
& Maximum archive size. Set to $1.5 \times$ the target archive size unless explicitly overridden. \\

Initial diversity $\tau$ 
& $0.10$ 
& Initial minimum distance used by the unstructured archive to decide whether a candidate is behaviorally distinct enough to be retained. This threshold is later adapted online as the archive evolves. Larger values enforce more separation; smaller values allow denser archives. \\

Nº of illumination itrs. 
& $1000$ 
& Total nº of candidate policies proposed and evaluated during MAP-Elites search. Higher values allow broader exploration but increase compute cost. Each iteration samples a parent, mutates it, evaluates the child, and attempts archive insertion. \\

Mutation scale 
& $0.05$ 
& Strength of parameter-space mutation applied to sampled parents. Larger values encourage broader exploration, while smaller values keep mutations more local and conservative. \\

Evaluation episodes
& $50$
& Number of episodes used to estimate each candidate's return, success rate, and behavior descriptor before archive insertion. \\

\bottomrule
\end{tabular}
\caption{MAP-Elites hyperparameters used in the illumination stage.}
\label{tab:map_elites_hparams}
\vspace{-2.0em}
\end{table*}

\paragraph{MAP-Elites Algorithm}
Algorithm~\ref{alg:map_elites_hpvd} summarizes the whole illumination MAP-Elites procedure used in \textsc{TeLAPA}.
In the implementation, MAP-Elites operates as a parameter-space illumination procedure that initializes the archive with the converged base policy, repeatedly samples a parent from the archive, mutates it, evaluates the resulting candidate over several episodes, and inserts accepted candidates while adapting the archive diversity threshold online.
The mutation step is deliberately coupled to a competence gate: candidate perturbations are not stored because they are different, but because they remain successful on the source task while satisfying the archive’s behavioral spacing criterion.

\begin{algorithm}[ht]
\caption{MAP-Elites for task-local archive construction}
\label{alg:map_elites_hpvd}
\small
\KwIn{
environment $e_t$; reference policy $\theta_t$; base tag $\bar{\tau}_t$;\\
shared embedder $f_{\phi}$; current embedding state $\mathcal{E}_{t-1}$;\\
illumination budget $N_{\mathrm{ME}}$; evaluation episodes $m$; initial mutation scale $\sigma_0$;\\
optional initial archive $\mathcal{H}^{(0)}$; optional injection pool $\mathcal{P}$
}
\KwOut{
task archive $\mathcal{H}_{\bar{\tau}_t}$
}

\tcc{1. Archive initialization}
\uIf{$\mathcal{H}^{(0)}$ is provided}{
    initialize $\mathcal{H}_{\bar{\tau}_t}\gets \mathcal{H}^{(0)}$\;
}
\Else{
    create an empty unstructured archive $\mathcal{H}_{\bar{\tau}_t}$ with target size and spacing $d_{\min}$\;
}

evaluate $\theta_t$ on $e_t$ for $m$ episodes and collect trajectories $\Xi^{\mathrm{base}}$\;
compute raw latent statistics from $\Xi^{\mathrm{base}}$\;
encode them under $\mathcal{E}_{t-1}$ to obtain descriptor $z^{\mathrm{bd}}_{\mathrm{base}}$\;
insert the base elite $(z^{\mathrm{bd}}_{\mathrm{base}}, \theta_t, F_{\mathrm{base}}, \mathrm{SR}_{\mathrm{base}}, \sigma_0)$ into $\mathcal{H}_{\bar{\tau}_t}$\;
store the archive reference descriptor $z^{\mathrm{ref}}_{\bar{\tau}_t}\gets z^{\mathrm{bd}}_{\mathrm{base}}$\;
\BlankLine

\tcc{2. Illumination loop}
\For{$i=1,\dots,N_{\mathrm{ME}}$}{
    choose a parent elite $x_p$ from $\mathcal{H}_{\bar{\tau}_t}$\;
    
    \If{$\mathcal{P}\neq\varnothing$ and elite injection is triggered}{
        optionally replace $x_p$ with a candidate from $\mathcal{P}$ to improve coverage of underrepresented regions\;
    }

    self-adapt the mutation scale from the parent's scale to obtain $\sigma_i$\;
    mutate the parent parameters to obtain candidate policy $\tilde{\theta}_i$\;

    evaluate $\tilde{\theta}_i$ on $e_t$ for $m$ episodes and collect trajectories $\Xi_i$\;
    compute raw latent statistics from $\Xi_i$\;
    encode them under $\mathcal{E}_{t-1}$ to obtain descriptor $z^{\mathrm{bd}}_i$\;
    compute the competence threshold $g_i$\;

    \uIf{$\mathrm{SR}(\tilde{\theta}_i)\ge g_i$}{
        insert elite $(z^{\mathrm{bd}}_i,\tilde{\theta}_i,F_i,\mathrm{SR}_i,\sigma_i)$ into $\mathcal{H}_{\bar{\tau}_t}$\;
        \If{the archive changed}{
            adapt the spacing threshold $d_{\min}$\;
        }
    }
    save archive snapshots and summary statistics\;
}

restore the original reference policy $\theta_t$\;
\Return{$\mathcal{H}_{\bar{\tau}_t}$}\;
\end{algorithm}
\vspace{-1.0em}
\section{TeLAPA}
\label{app:telapa_hp_algs}

\paragraph{TeLAPA Hyperparameters.} 
Tables~\ref{tab:hp_search_telapa} summarize the explored grids and the final selected configuration used in the main experiments for the online latent space maintenance procedure described in App.~\ref{app:learned_embedding_v2}.
We also provide a description of what each hyperparameter controls:
\begin{itemize}
    \item \textbf{Embedding Steps}: Number of gradient steps to train the trajectory embedder at each task boundary
    \item \textbf{Embedding Learning Rate}: Learning rate for the trajectory embedder boundary update using the Adam optimizer.
    \item \textbf{Anchor Bank Fraction}: Fraction of normalizer refit bank drawn from anchors (rest from replay).
    \item \textbf{Store as Anchor Threshold}: SR threshold for adding episode-sets to anchor bank.
    \item \textbf{Min. Bank Sets}: Minimum number of episode-sets available in anchors+replay required to run the boundary embedder update.
    \item \textbf{$w_{\mathrm{contrast}}$}: Weight on the contrastive InfoNCE term.
    \item \textbf{$w_{\mathrm{distill}}$}: Weight for teacher distillation loss on anchors (anti-drift) during boundary training.
    \item \textbf{$\lambda_{\mathrm{norm}}$}: Weight for the \emph{scale} component of the anchor distillation loss (norm-matching). Distillation already matches teacher vs student \emph{direction} via MSE on normalized embeddings; this term additionally matches $\|z_{\mathrm{student}}\|_2$ to $\|z_{\mathrm{teacher}}\|_2$ on the anchor subset to reduce embedding inflation or collapse.
    \item \textbf{Embedding Temperature $\tau$}: InfoNCE temperature for the contrastive loss in embedder training.
\end{itemize}

\begin{table*}[ht]
\centering
\begin{tabular}{llc}
\toprule
\textbf{Hyperparameter} & \textbf{Explored values} & \textbf{Selected value} \\
\midrule
Embedding Steps            & $\{300,\;500,\;700\}$                    & $700$ \\
Embedding Learning Rate    & $\{10^{-3},\;2\!\times\!10^{-4},\;3\!\times\!10^{-4},\;5\!\times\!10^{-4}\}$ & $5\times 10^{-4}$ \\
Anchor Bank Fraction           & $\{0.20,\;0.33,\;0.50\}$                 & $0.33$ \\
Store as Anchor Threshold      & $\{0.50,\;0.60,\;0.70\}$                 & $0.50$ \\
Min. Bank Sets                 & $\{12,\;24,\;32\}$                       & $32$ \\
$w_{\mathrm{contrast}}$        & $\{0.5,\;1.0,\;2.0\}$                         & $1.0$ \\
$w_{\mathrm{distill}}$         & $\{0.5,\;1.0,\;2.0\}$                  & $1.0$ \\
$\lambda_{\mathrm{norm}}$      & $\{0.0,\;0.25,\;0.5,\;1.0\}$           & $1.0$ \\
Embedding Temperature $\tau$   & $\{0.07,\;0.10,\;0.15,\;0.22\}$        & $0.15$ \\
\bottomrule
\end{tabular}
\vspace{-0.5em}
\caption{Hyperparameter search space explored for \textsc{TeLAPA}, together with the final selected configuration used in the main experiments.}
\label{tab:hp_search_telapa}
\end{table*}

\vspace{-1.0em}
\subsection{TeLAPA Algorithm}
\label{app:alf_telapa}

Algorithm~\ref{alg:hpvd_overall} summarizes the full TeLAPA pipeline at the sequence level.
The method alternates between: (i) archive-based initialization for the incoming task, 
(ii) PPO adaptation while collecting trajectory evidence, (iii) archive illumination or revisit update,
and (iv) boundary maintenance of the shared latent space.
For details about MAP-Elites see Appendix \ref{app:map_elites}. For details about the components of the online re-embedding procedure check Appendix \ref{app:learned_embedding_v2}.

\begin{algorithm}[ht]
\small
\DontPrintSemicolon
\SetAlgoLined
\SetKwInOut{KwIn}{Input}
\SetKwInOut{KwOut}{Output}

\SetKwFunction{FewShotSelectElite}{FewShotSelectElite}
\SetKwFunction{TrainTask}{TrainTask}
\SetKwFunction{BuildArchive}{BuildArchive}
\SetKwFunction{RefreshArchive}{RefreshArchive}
\SetKwFunction{Maintain}{BoundaryMaintenance}
\SetKwFunction{BaseTag}{BaseTag}

\KwIn{
    task sequence $\mathcal{S}=\{(\tau_t,e_t)\}_{t=1}^{T}$ \\
    shared trajectory embedder $f_\phi$ \\
    PPO budget $B_{\mathrm{ppo}}$ \\
    illumination budget $B_{\mathrm{ME}}$
}
\KwOut{
    trained policy sequence $\{\theta_t\}_{t=1}^{T}$ and archive set $\mathcal{H}$
}
\BlankLine

Initialize archive dictionary $\mathcal{H}\leftarrow \varnothing$\;
Initialize embedding manager $\mathcal{M}\leftarrow (E_0,\texttt{no normalizer})$\;
Initialize anchor store $\mathcal{A}$ and replay store $\mathcal{R}$\;
Initialize previous policy snapshot $\theta_{\mathrm{prev}}\leftarrow \varnothing$\;
\BlankLine

\For{$t=1,\dots,T$}{
    \tcc{1. Archive-based Task Initialization \& Policy Seeding}
    Build a fresh PPO learner for environment $e_t$\;
    $b_t \leftarrow$ \BaseTag{$\tau_t$} \tcp*[r]{drop prime marks on revisits}
    \BlankLine
    \uIf{$\mathcal{H}\neq \varnothing$}{
        $\theta^{(0)}_t \leftarrow$ \FewShotSelectElite{$\tau_t, \mathcal{H}, \mathcal{M}$}\;
        Initialize PPO with $\theta^{(0)}_t$\;
    }
    \Else{
        Initialize PPO from scratch\;
    }
    \BlankLine

    \tcc{2. Task Training \& Evaluation}
    \textbf{Pre-eval:} Evaluate initial policy on $e_t$\;
    \textbf{Elite Storage:} Add resulting trajectories to $\mathcal{R}$ and high-success/diverse ones to $\mathcal{A}$\;
    \textbf{Optimize:} $\theta_t \leftarrow$ \TrainTask{$e_t, \theta^{(0)}_t, B_{\mathrm{ppo}}, \mathcal{A}, \mathcal{R}$}\;
    \textbf{Post-eval:} Evaluate final policy $\theta_t$ on $e_t$\;
    \textbf{Final Storage:} Append final trajectories to $\mathcal{R}$, and successful ones to $\mathcal{A}$\;
    \BlankLine

    \tcc{3. MAP-Elites (Illumination \& Archive Update)}
    \uIf{$b_t \notin \mathcal{H}$}{
        $\mathcal{H}[b_t] \leftarrow$ \BuildArchive{$e_t, \theta_t, B_{\mathrm{ME}}, f_\phi, \mathcal{M}$}\;
    }
    \Else{
        $\mathcal{H}[b_t] \leftarrow$ \RefreshArchive{$\mathcal{H}[b_t], \theta_t, e_t, f_\phi, \mathcal{M}$}\;
    }
    \BlankLine

    \tcc{4. Online Latent Space Maintenance}
    $\mathcal{M},\mathcal{H} \leftarrow$ \Maintain{$f_\phi, \mathcal{M}, \mathcal{A}, \mathcal{R}, \mathcal{H}$}\;
    $\theta_{\mathrm{prev}}\leftarrow \theta_t$\;
}

\caption{TeLAPA: Continual reinforcement learning with archive-based transfer and online latent-space maintenance.}
\label{alg:hpvd_overall}
\end{algorithm}
\vspace{-10pt}

\subsubsection{\textsc{Few-Shot}: Transfer-Origin Selection}
\label{app:fewshot_select_elite}

Algorithm~\ref{alg:fewshot_select_elite} defines the transfer-origin selection routine used in Algorithm~\ref{alg:hpvd_overall}. 
Given the incoming task, the archive library, and the current embedding manager, the routine first loads the available archives and gathers a compact set of candidate elites using the current archive geometry. 
Then, each candidate elite $\eta_j$ is tested on the incoming task by loading its archived weights into the PPO learner, resetting optimizer to avoid inherited optimizer momentum, and rollout state, and running a short adaptation probe. 
The probe returns an initial success rate $\mathrm{SR}^{0}_j$, a final success rate $\mathrm{SR}^{\mathrm{final}}_j$, a success-rate slope $\mathrm{slope}_j$, and a probe area-under-curve $\mathrm{AUC}_j$. 
Here, $\mathrm{slope}_j$ measures short-horizon improvement during the probe, while $\mathrm{AUC}_j$ measures average success across the probe trajectory. 
Selection is conservative: we first identify the best final probe success rate $\mathrm{SR}^{\star}=\max_j \mathrm{SR}^{\mathrm{final}}_j$ and keep only candidates within an $\epsilon$-window of this value. 
Within this near-best set, \textsc{Score} ranks candidates using a weighted combination of normalized final success, recovery slope, and probe AUC. 
This prevents selecting candidates that improve quickly but remain weak, while still favoring fast-recovering candidates among those with comparable final performance. 
The probe is used only for selection; the returned initialization is the original archived elite, not the post-probe weights.

\begin{algorithm}[ht]
\small
\DontPrintSemicolon
\SetAlgoLined
\SetKwInOut{KwIn}{Input}
\SetKwInOut{KwOut}{Output}

\SetKwFunction{LoadArchives}{LoadArchives}
\SetKwFunction{GatherCandidates}{GatherCandidates}
\SetKwFunction{Probe}{Probe}
\SetKwFunction{Score}{Score}

\KwIn{
    incoming task environment $e_{k+1}$ \\
    archive paths $\mathcal{P}$ \\
    PPO learner $M$ \\
    embedding manager $\mathcal{M}=(\phi_t,\mathcal{N}_t)$ \\
    candidate budget $K_{\mathrm{pool}}$ \\
    probe budget $B_{\mathrm{probe}}$ \\
    final-success window $\epsilon$
}
\KwOut{
    selected initialization $\theta^{(0)}_{k+1}$
}
\BlankLine

$\mathcal{H} \leftarrow$ \LoadArchives{$\mathcal{P}$}\;
$\mathcal{C}_{k+1} \leftarrow$ \GatherCandidates{$\mathcal{H}, K_{\mathrm{pool}}, \mathcal{M}$}\;
\BlankLine

\ForEach{$\eta_j \in \mathcal{C}_{k+1}$}{
    Load archived weights $\theta_{\eta_j}$ into $M$\;
    Reset optimizer and rollout state of $M$\;
    $(\mathrm{SR}^{0}_j,\mathrm{SR}^{\mathrm{final}}_j,\mathrm{slope}_j,\mathrm{AUC}_j) 
    \leftarrow$ \Probe{$M,e_{k+1},B_{\mathrm{probe}}$}\;
}
\BlankLine

$\mathrm{SR}^{\star} \leftarrow \max_j \mathrm{SR}^{\mathrm{final}}_j$\;
$\mathcal{I} \leftarrow \{j : \mathrm{SR}^{\mathrm{final}}_j \geq \mathrm{SR}^{\star}-\epsilon\}$\;
\BlankLine

$j^\star \leftarrow 
\arg\max_{j\in\mathcal{I}}$
\Score{$\mathrm{SR}^{\mathrm{final}}_j,\mathrm{slope}_j,\mathrm{AUC}_j,\mathrm{SR}^{\star}$}\;
\BlankLine

\Return original archived weights $\theta_{\eta_{j^\star}}$ as $\theta^{(0)}_{k+1}$\;

\caption{\textsc{FewShotSelectElite}: few-shot transfer-origin selection.}
\label{alg:fewshot_select_elite}
\end{algorithm}

\section{Learned trajectory embedding under drift}
\label{app:learned_embedding_v2}

This appendix details the learned behavioral embedding used during the online re-embedding under drift.
Unlike in any fixed-encoder variant, the embedding space is not static: the trajectory encoder is updated at task boundaries using anchor and replay data, and previously stored archives are periodically re-expressed under the current latent geometry.

\vspace{-0.5em}
\subsection{Algorithm: Online latent-space maintenance}
Algorithm~\ref{alg:latent_maintenance} summarizes the boundary maintenance step used in TeLAPA.
After each task, the shared trajectory embedder is updated from accumulated anchor and replay trajectories, the global latent normalizer is refit under the updated encoder, and all previously saved archives are re-expressed under the current geometry.

\begin{algorithm}[ht]
\caption{Online latent-space maintenance}
\label{alg:latent_maintenance}
\small
\KwIn{
shared embedder $f_{\phi}$; embedding state $\mathcal{E}_{t-1}$; anchor buffer $\mathcal{A}$; replay buffer $\mathcal{R}$;\\
archive set $\{\mathcal{H}_j\}$; minimum bank size $B_{\min}$; training steps $K$
}
\KwOut{
updated embedding state $\mathcal{E}_t$ and maintained archive set $\{\mathcal{H}_j\}$
}

\tcc{1. Check whether boundary maintenance is possible}
form the bank $\mathcal{B} \gets \mathcal{A} \cup \mathcal{R}$\;
\uIf{$|\mathcal{B}| < B_{\min}$}{
    $\mathcal{E}_t \gets \mathcal{E}_{t-1}$\;
    \Return{$(\mathcal{E}_t,\{\mathcal{H}_j\})$}\;
}
take a frozen teacher snapshot $\tilde{f}_{\phi} \gets f_{\phi}$\;
\BlankLine

\tcc{2. Update the embedder on anchor/replay batches}
\For{$k=1,\dots,K$}{
    sample a mixed mini-batch from $\mathcal{A}$ and $\mathcal{R}$\;
    compute a contrastive objective on augmented trajectory views\;
    \If{enough anchor samples are available}{
        add a teacher-student distillation term on anchor examples only\;
    }
    update $f_{\phi}$ by one gradient step\;
}
\BlankLine

\tcc{3. Refit the latent normalizer}
bump the embedding version\;
sample a representative bank from $\mathcal{A}$ and $\mathcal{R}$\;
refit the global latent normalizer under the updated embedder\;
set $\mathcal{E}_t$ to the new embedder-normalizer state\;
\BlankLine

\tcc{4. Re-express all archives under the current geometry}
\ForEach{archive $\mathcal{H}_j$ in $\{\mathcal{H}_j\}$}{
    save a stale pre-maintenance snapshot of $\mathcal{H}_j$\;
    
    re-embed archive elites from stored trajectory sketches under $\mathcal{E}_t$\;
    
    \If{too many elites lack stored trajectory information}{
        re-evaluate those elites on their native task to rebuild trajectory sketches and re-embed them\;
    }
    
    refresh the archive reference descriptor in the same updated latent space\;
    repack the archive under the current geometry to enforce spacing and capacity constraints\;
    stamp archive metadata with the new embedding version and normalizer identifier\;
    save the maintained archive\;
}
\Return{$(\mathcal{E}_t,\{\mathcal{H}_j\})$}\;
\end{algorithm}

\vspace{-1.5em}
\subsection{Trajectory features, episodes, and episode sets}
\label{app:traj_features}

Behavior is logged through a trajectory wrapper that records a compact symbolic feature vector at every environment step.
For a rollout of length $T$, the episode is represented as
\begin{equation}
E = [f_1,\dots,f_T] \in \mathbb{R}^{T \times D},
\qquad D=11,
\end{equation}
where each timestep feature vector is
\begin{equation}
\begin{split}
f_t &= \bigl[ x_t^{01},\, y_t^{01},\, d_t^{01},\, \tau_t^{01},\, a_t^{01},\, \{\texttt{has\_key}_t\}, \\
    &\quad\, \{\texttt{has\_ball}_t\},\, \{\texttt{has\_box}_t\},\, \{\texttt{door\_open}_t\},\, \{\texttt{box\_toggled}_t\},\, \{\texttt{delivered}_t\} \bigr].
\end{split}
\label{eq:traj_feat_vector}
\end{equation}
The first five coordinates are normalized geometric and control features:
\begin{align}
x_t^{01} &= \mathrm{clip}\!\left(\frac{x_t}{\max(W-1,1)}, 0, 1\right),\\
y_t^{01} &= \mathrm{clip}\!\left(\frac{y_t}{\max(H-1,1)}, 0, 1\right),\\
d_t^{01} &= \mathrm{clip}\!\left(\frac{d_t}{3}, 0, 1\right),\\
\tau_t^{01} &= \mathrm{clip}\!\left(\frac{\texttt{step\_count}_t}{\texttt{max\_steps}}, 0, 1\right),\\
a_t^{01} &=
\begin{cases}
\mathrm{clip}\!\left(\dfrac{a_t}{n_{\mathrm{act}}-1}, 0, 1\right), & n_{\mathrm{act}} > 1,\\[6pt]
a_t, & \text{otherwise},
\end{cases}
\end{align}
where $(x_t,y_t)$ is the agent position, $d_t \in \{0,1,2,3\}$ is the agent direction, and $a_t$ is the executed action.
The remaining six coordinates are sticky binary task-event indicators provided by the upstream event wrapper.
Thus, the encoder does not consume raw observations directly; it consumes a sequence of compact per-step behavioral features.

During boundary updates, data are sampled not as isolated episodes but as \emph{episode sets}.
An episode set is a list of rollout episodes associated with the same stored unit in anchor or replay memory:
\begin{equation}
\mathcal{S}_i = \{E_{i,1},\dots,E_{i,n_i}\},
\end{equation}
Anchor and replay buffers store collections of episode sets rather than single trajectories.
Given a batch of $B$ episode sets,
\begin{equation}
\mathcal{B} = \{\mathcal{S}_1,\dots,\mathcal{S}_B\},
\end{equation}
the training step samples one episode uniformly from each set,
\begin{equation}
\tilde E_i \sim \mathcal{U}(\mathcal{S}_i), 
\end{equation}
pads or truncates the resulting episodes to a common temporal cap $T_{\max}$, and forms
\begin{equation}
X \in \mathbb{R}^{B \times T_{\max} \times D},
\qquad
\ell \in \{0,\dots,T_{\max}\}^{B},
\end{equation}
where $\ell_i$ is the valid length of the $i$-th sampled episode.
In the current implementation, $T_{\max}=256$.

\subsection{Episode encoder architecture}
\label{app:encoder_architecture}

The learned trajectory encoder maps an episode $E \in \mathbb{R}^{T \times D}$ to a latent vector with latent dimension $d=8$:
\begin{equation}
z_e \in \mathbb{R}^{d},
\end{equation}
The encoder consists of three stages.

\paragraph{Per-step MLP.}
Each per-step feature vector is first transformed independently:
\begin{equation}
h_t^{(0)}
=
\mathrm{MLP}_{\mathrm{step}}(f_t)
=
\varphi\!\left(
W_2 \, \varphi(W_1 f_t + b_1) + b_2
\right),
\end{equation}
where $\varphi(\cdot)$ is a ReLU nonlinearity.
This produces a sequence
\begin{equation}
H^{(0)} = [h_1^{(0)},\dots,h_T^{(0)}].
\end{equation}

\paragraph{Temporal aggregation with a GRU.}
The per-step embeddings are then processed by a recurrent aggregator:
\begin{equation}
h_t = \mathrm{GRU}(h_{t-1}, h_t^{(0)}),
\end{equation}
and the final hidden state is used as the episode summary:
\begin{equation}
h_E = h_T.
\end{equation}
The GRU is used to preserve temporal ordering and to summarize variable-length trajectories into a fixed-size representation.

\paragraph{Projection to the latent descriptor.}
The episode-level latent is obtained via a projection head:
\begin{equation}
z_e
=
\mathrm{Proj}(h_E)
=
W_4 \, \varphi(W_3 h_E + b_3) + b_4
\in \mathbb{R}^{d}.
\end{equation}

Thus, the full episode encoder is
\begin{equation}
z_e = \phi(E),
\end{equation}
where $\phi$ denotes the current shared trajectory embedder.


\subsection{Policy-level latent summaries}
\label{app:policy_level_summary}

A policy is evaluated over a collection of episodes
\begin{equation}
\mathcal{E}(\pi)=\{E^{(1)},\dots,E^{(M)}\}.
\end{equation}
The encoder is applied to every episode:
\begin{equation}
z_e^{(m)} = \phi(E^{(m)}), \qquad m=1,\dots,M.
\end{equation}

The policy-level mean descriptor is
\begin{equation}
z_{\mathrm{mean}}(\pi)
=
\frac{1}{M}\sum_{m=1}^{M} z_e^{(m)}.
\end{equation}
This is the principal behavioral descriptor used for archive geometry.

We also store two variability summaries.

\paragraph{Across-episode variability.}
The first is the standard deviation across episode embeddings:
\begin{equation}
z_{\mathrm{std,ep}}(\pi)
=
\sqrt{
\frac{1}{M}
\sum_{m=1}^{M}
\left(z_e^{(m)} - z_{\mathrm{mean}}(\pi)\right)^2
},
\end{equation}
where the square root is taken element-wise.

\paragraph{Within-episode temporal variability.}
To quantify variation over time inside an episode, we also compute per-timestep latent projections.
Let
\begin{equation}
z^{(m)}_{1:T_m}
=
\bigl[z^{(m)}_1,\dots,z^{(m)}_{T_m}\bigr]
\in \mathbb{R}^{T_m \times d}
\end{equation}
denote the sequence of per-timestep projected embeddings for episode $m$.
We define the within-episode standard deviation
\begin{equation}
s^{(m)}_{\mathrm{time}}
=
\sqrt{
\frac{1}{T_m}
\sum_{t=1}^{T_m}
\left(z^{(m)}_t - \bar z^{(m)}\right)^2
},
\qquad
\bar z^{(m)} = \frac{1}{T_m}\sum_{t=1}^{T_m} z^{(m)}_t,
\end{equation}
and then average across episodes:
\begin{equation}
z_{\mathrm{std,time}}(\pi)
=
\frac{1}{M}
\sum_{m=1}^{M}
s^{(m)}_{\mathrm{time}}.
\end{equation}

The tuple
\begin{equation}
\Bigl(
z_{\mathrm{mean}}(\pi),
z_{\mathrm{std,ep}}(\pi),
z_{\mathrm{std,time}}(\pi)
\Bigr)
\end{equation}
is the full raw latent summary attached to a policy evaluation.

\subsection{View generation for contrastive training}
\label{app:contrastive_views}

Boundary training uses two augmented views of the same sampled episode.
Given an episode
\begin{equation}
E = [f_1,\dots,f_T],
\end{equation}
we generate two stochastic views
\begin{equation}
\widetilde E^{(1)}, \widetilde E^{(2)}.
\end{equation}
Each view is produced by the following transformations.

\paragraph{Temporal crop.}
A contiguous sub-trajectory is sampled by choosing a crop length
\begin{equation}
L' \sim \mathcal{U}\bigl(\{\lfloor \rho T \rfloor,\dots,T\}\bigr),
\end{equation}
where $\rho \in (0,1]$ is the minimum crop fraction, and then selecting a start index $s$ uniformly from the valid range.
The cropped episode is
\begin{equation}
E_{\mathrm{crop}} = [f_s,\dots,f_{s+L'-1}].
\end{equation}
In the current implementation, $\rho=0.6$ by default.

\paragraph{Feature dropout.}
A binary mask
\begin{equation}
m \in \{0,1\}^{D}
\end{equation}
is sampled once per episode view and then applied uniformly across time:
\begin{equation}
\widetilde f_t = m \odot f_t.
\end{equation}
This removes a subset of feature channels while keeping the temporal structure intact.
The same mask is used at every timestep of that view.

\paragraph{Additive Gaussian noise.}
Finally, small independent Gaussian noise is added:
\begin{equation}
\widetilde f_t
\leftarrow
\widetilde f_t + \sigma_{\mathrm{noise}} \, \varepsilon_t,
\qquad
\varepsilon_t \sim \mathcal{N}(0, I_D).
\end{equation}

These augmentations encourage the encoder to represent stable behavioral structure rather than overfit to exact timestep-level realizations.

\subsection{Teacher--student boundary update}
\label{app:teacher_student_update}

After task $t$, let $\phi_t$ denote the current shared embedder.
If the anchor and replay stores contain enough episode sets, we perform a boundary update.
A frozen copy of the pre-update encoder is created,
\begin{equation}
\tilde \phi_t \leftarrow \phi_t,
\end{equation}
and used as the \emph{teacher}.
The live encoder $\phi_t$ becomes the \emph{student} and is optimized for a fixed number of steps on mixed anchor--replay batches.

\paragraph{Batch construction.}
Each boundary step samples a mixed batch of episode sets from the anchor store and replay store:
\begin{equation}
\mathcal{B} = \{\mathcal{S}_1,\dots,\mathcal{S}_B\},
\end{equation}
with a prescribed anchor fraction.
A minimum number of anchor examples may also be enforced per batch.
From each set, one episode is sampled and converted into two augmented views as described above.

\paragraph{Student embeddings for the two views.}
Let
\begin{equation}
z^{(1)}_i = \phi_t\!\left(\widetilde E^{(1)}_i\right),
\qquad
z^{(2)}_i = \phi_t\!\left(\widetilde E^{(2)}_i\right),
\end{equation}
be the student embeddings for the two augmented views of the $i$-th sampled episode.

\subsubsection{Contrastive loss}
\label{app:contrastive_loss}

The contrastive term is a symmetric InfoNCE objective.
First normalize the view embeddings:
\begin{equation}
\bar z^{(1)}_i = \frac{z^{(1)}_i}{\|z^{(1)}_i\|_2},
\qquad
\bar z^{(2)}_i = \frac{z^{(2)}_i}{\|z^{(2)}_i\|_2}.
\end{equation}
Then define the similarity matrix
\begin{equation}
s_{ij}
=
\frac{
\left(\bar z^{(1)}_i\right)^\top \bar z^{(2)}_j
}{
\tau
},
\end{equation}
where $\tau>0$ is a temperature hyperparameter. In our implementation $\tau=0.1$.

The forward and backward cross-entropy terms are
\begin{align}
\mathcal{L}_{1 \rightarrow 2}
&=
-\frac{1}{B}\sum_{i=1}^{B}
\log
\frac{\exp(s_{ii})}{\sum_{j=1}^{B}\exp(s_{ij})},\\
\mathcal{L}_{2 \rightarrow 1}
&=
-\frac{1}{B}\sum_{i=1}^{B}
\log
\frac{\exp(s_{ii})}{\sum_{j=1}^{B}\exp(s_{ji})}.
\end{align}
The final contrastive loss is
\begin{equation}
\mathcal{L}_{\mathrm{contrast}}
=
\frac{1}{2}
\left(
\mathcal{L}_{1 \rightarrow 2}
+
\mathcal{L}_{2 \rightarrow 1}
\right).
\end{equation}

This encourages two perturbed views of the same episode to map nearby while different batch items act as negatives.




\subsubsection{Anchor-masked teacher distillation}
\label{app:distill_loss} 

The teacher gives us a target embedding for each anchor episode. We want the student to stay close to that target in a smooth, stable way.
Therefore, the purpose of distillation is to constrain representation drift in trusted reference behaviors.
Unlike the contrastive branch, teacher--student matching is computed on the \emph{original, non-augmented} sampled episodes.
Let
\begin{equation}
z^{\mathrm{teach}}_i = \tilde \phi_t(E_i),
\qquad
z^{\mathrm{stud}}_i = \phi_t(E_i).
\end{equation}
Let $\mathcal{I}_{\mathrm{anc}} \subseteq \{1,\dots,B\}$ denote the anchor subset of the batch.
Distillation is applied only when enough anchor data are available globally.

We use an MSE-based distillation loss to provide a direct regression to the teacher representation. It consists of a directional term on $L_2$-normalized embeddings that aims to keep the teacher and student pointing in the same latent direction:
\begin{equation}
\mathcal{L}_{\mathrm{distill}}
=
\mathrm{MSE}\!\left(
\mathrm{norm}(z^{\mathrm{stud}}_i),
\mathrm{norm}(z^{\mathrm{teach}}_i)
\right)_{i\in\mathcal{I}_{\mathrm{anc}}}
\end{equation}
However, once we normalize the embeddings, we lose the magnitude information. Thus, the student could preserve the direction while letting the latent space inflate or collapse in norm. Therefore, to preserve not just the direction, but also the approximate scale, we also match embedding magnitudes:
\begin{equation}
\mathrm{MSE}\!\left(
\|z^{\mathrm{stud}}_i\|_2,
\|z^{\mathrm{teach}}_i\|_2
\right)_{i\in\mathcal{I}_{\mathrm{anc}}}.
\end{equation}
The distillation term is
\begin{equation}
\mathcal{L}_{\mathrm{distill}}
=
\mathcal{L}_{\mathrm{dir}}
+
\lambda_{\mathrm{norm}}
\mathcal{L}_{\mathrm{scale}}.
\end{equation}

\subsubsection{Total boundary objective}
\label{app:total_emb_loss}

The student encoder is trained with the composite objective\footnote{Training-time losses live in raw encoder space, while archive-time geometry lives in globally normalized latent-space.}
\begin{equation}
\mathcal{L}_{\mathrm{emb}}
=
w_{\mathrm{contrast}} \mathcal{L}_{\mathrm{contrast}}
+
w_{\mathrm{distill}} \mathcal{L}_{\mathrm{distill}}.
\end{equation}
Only the student parameters are updated. The teacher remains frozen throughout the boundary step.

\subsection{Global latent normalization}
\label{app:global_normalizer}

The encoder produces raw policy-level latent summaries, but archive geometry is defined in a shared normalized space.
To maintain a single within-run coordinate system, the global latent normalizer maps raw policy-level latent means into the shared behavior-descriptor space used for archive geometry.
Thus, the normalizer is fit \emph{globally} after each boundary update from a bank of anchor and replay \emph{episode sets}, not separately per archive, and then reused for elite insertion, archive re-embedding, and all subsequent distance-based operations.

Let
\begin{equation}
\mathfrak{B}_t = \{\mathcal{S}_1,\dots,\mathcal{S}_N\}
\end{equation}
denote the bank used at boundary $t$, where each $\mathcal{S}_i$ is a list of episodes.
For each episode set, we compute a raw policy-level mean descriptor via the current encoder:
\begin{equation}
z_i = z_{\mathrm{mean}}(\mathcal{S}_i) \in \mathbb{R}^{d}.
\end{equation}
Empty episode sets, malformed outputs, and non-finite descriptors are discarded.
Stacking the remaining descriptors yields
\begin{equation}
Z_t =
\begin{bmatrix}
z_1^\top\\
\vdots\\
z_N^\top
\end{bmatrix}
\in \mathbb{R}^{N \times d}.
\end{equation}

The global normalizer is then fit component-wise using robust statistics:
\begin{align}
\mu_t &= \mathrm{median}(Z_t) \in \mathbb{R}^{d},\\
q_{25,t} &= Q_{25}(Z_t) \in \mathbb{R}^{d},\\
q_{75,t} &= Q_{75}(Z_t) \in \mathbb{R}^{d},\\
\sigma_t &= \max\!\left(\frac{q_{75,t} - q_{25,t}}{1.349},\; \sigma_{\min}\right),
\end{align}
where all statistics are taken element-wise. We use $\sigma_{\min}=10^{-3}$ in the current implementation.
The factor $1.349$ converts IQR to the corresponding standard deviation under normality.
Given any raw descriptor $z \in \mathbb{R}^{d}$, the normalized descriptor is
\begin{equation}
\hat z
=
\mathcal{N}_t(z)
=
\frac{z - \mu_t}{\sigma_t + \epsilon},
\qquad
\epsilon = 10^{-8}.
\label{eq:global_normalizer}
\end{equation}

The descriptor used for archive geometry is therefore
\begin{equation}
z_{\mathrm{bd}}(\pi) = \mathcal{N}_t\!\left(z_{\mathrm{mean}}(\pi)\right).
\end{equation}
By default, only the policy-level mean descriptor is normalized in this way.
Thus, all nearest-neighbor queries, novelty distances, and archive insertions at boundary $t$ are performed using the shared transformation $\mathcal{N}_t$ applied to policy-level mean descriptors.

\section{Forgetting and Plasticity Through BWT Metrics}
\label{app:metrics_transfer}

Absolute backward transfer (BWT) is widely used to quantify forgetting, but in reinforcement learning it can be misleading due to \emph{floor effects}: a method that fails to learn early tasks (low post-training success) cannot ``forget'' much, producing deceptively small-magnitude BWT. 


\paragraph{Notation.}
Let $\mathrm{SR}_{\text{post}}(T)$ denote the success rate on task $T$ immediately after training on $T$ (i.e., post-task performance), and let $\mathrm{SR}_{\text{end}}(T)$ denote the success rate on $T$ at the end of the entire sequence. 

\paragraph{(A) Absolute BWT (SR units).}
We compute absolute forgetting as
\[
\mathrm{BWT} \;=\; \frac{1}{|\mathcal{T}_{\mathrm{prev}}|}\sum_{T\in\mathcal{T}_{\mathrm{prev}}}\Big(\mathrm{SR}_{\text{end}}(T)-\mathrm{SR}_{\text{post}}(T)\Big),
\]
where $\mathcal{T}_{\mathrm{prev}}$ is the set of previously encountered (non-current) tasks. Values near $0$ indicate little net change; more negative values indicate forgetting. However, BWT alone cannot distinguish ``retained knowledge'' from ``knowledge never acquired''.

\paragraph{(B) Coverage (learned-at-train-time).}
First, for each task, this target $\tau_T$ is defined as
\[
\tau_T = \max\!\bigl(0.9 \cdot \overline{\mathrm{bestSR}}_T,\ \tau_{\min}\bigr),
\]
where $\overline{\mathrm{bestSR}}_T$ is the mean best success rate achieved by task-specific \textsc{Scratch} runs on task $T$, and $\tau_{\min}$ is a fixed fallback threshold. This yields a rigorous task-specific ``baseline-to-beat'' standard while preventing trivially low targets on difficult tasks.

Then, to measure whether forgetting is meaningful, we report the fraction of tasks that were learned to the competence threshold when they were trained: 
\[
\mathrm{Coverage} \;=\; \Pr\big[\mathrm{SR}_{\text{post}}(T)\ge \tau_T\big].
\]
Low coverage implies the method rarely achieved competent performance, so absolute BWT is expected to be artificially small.

\paragraph{(C) Normalized BWT on learned tasks (nBWT).}
To evaluate forgetting \emph{conditional on having learned the task}, we normalize the performance drop and restrict to eligible tasks:
\[
\mathcal{E}=\{T:\mathrm{SR}_{\text{post}}(T)\ge\tau_T\},\qquad
\mathrm{nBWT}=\frac{1}{|\mathcal{E}|}\sum_{T\in\mathcal{E}}
\frac{\mathrm{SR}_{\text{end}}(T)-\mathrm{SR}_{\text{post}}(T)}{\max(\mathrm{SR}_{\text{post}}(T),\epsilon)}.
\]
This reports a relative retention loss on tasks that were actually learned. If $|\mathcal{E}|=0$, nBWT is undefined (NaN), which itself is informative: the method never achieved competence under $\tau_T$. Higher is better (i.e., closer to 0 or positive).
\begin{itemize}
    \item If nBWT = 0: no average change $\rightarrow$ method kept what it learned (good).
    \item If nBWT $<$ 0: performance dropped $\rightarrow$ forgetting (worse as it gets more negative).
    \item If nBWT $>$ 0: performance improved relative to post-training (can happen due to positive backward transfer / later training helping earlier tasks).
\end{itemize}


\paragraph{(D) Threshold retention (TR).}
We also report the end-of-sequence competence fraction:
\[
\mathrm{TR}\;=\;\Pr\big[\mathrm{SR}_{\text{end}}(T)\ge \tau_T\big].
\]
TR is easy to interpret and is robust to the floor-effect because it expresses ``how many tasks remain solved'' under a shared competence standard.


\paragraph{Interpretation.}
A method can appear to have ``low forgetting'' under absolute BWT simply because it never learned the tasks (low coverage), in which case nBWT is undefined or uninformative and TR is near zero. Conversely, methods with high coverage but low TR exhibit catastrophic forgetting: they learned tasks when trained but do not retain competence by the end. In our setting, the diagnostics clarify why absolute BWT must be contextualized and motivate the main text emphasis on revisit recoverability (TTT) and competence-based retention (TR).

\vspace{-0.5em}
\section{Additional Experimental Studies and Ablations}

\vspace{-0.5em}
\subsection{Task-Order Robustness: Alternative Curricula}
\label{app:curriculum_ablation}

Curriculum learning in reinforcement learning commonly improves learning by sequencing tasks or task distributions so that the agent first encounters easier or more informative settings before harder sparse-reward ones \citep{narvekar2020curriculum,florensa2017reverse}. 
However, in continual learning, task order itself can materially affect forgetting: Bell and Lawrence show that simply reordering the same task set can significantly change catastrophic forgetting \citep{bell2022effect}. 
More recent work further suggests that effective task orders need not coincide with a simple easy-to-hard progression, and may instead depend on broader structural relations among tasks \citep{li2025optimal}. 

Taking this into consideration, a natural concern arising from this paper is that our primary MiniGrid sequence was designed as a favorable hand-crafted curriculum, and that the gains of TeLAPA may depend critically on this ordering. 
Therefore, we include a task-order ablation that keeps the \emph{task set}, revisit protocol, and all training hyperparameters fixed, and varies \emph{only the order} in which tasks are presented.
See App.~\ref{app:envs} for an in-depth description of each task.

\paragraph{Primary curriculum (main text).}
Our main sequence follows a skill-ladder design:
\[
A \rightarrow B \rightarrow C \rightarrow D \rightarrow E \rightarrow A' \rightarrow B' \rightarrow C' \rightarrow D' \rightarrow E'.
\]
This ordering introduces new interaction requirements gradually, moving from navigation and text grounding to pickup, unlocking, clutter manipulation, and longer-horizon subgoal chaining.

\paragraph{Alternative curriculum 1: reverse anti-curriculum.}
The first alternative is a strict reverse ordering:
\[
E \rightarrow D \rightarrow C \rightarrow B \rightarrow A \rightarrow E' \rightarrow D' \rightarrow C' \rightarrow B' \rightarrow A'.
\]
This is the strongest anti-curriculum variant. 
It removes the easy-to-hard scaffold and instead presents the longest-horizon task first. 
If TeLAPA's advantage depends mainly on a favorable human-designed curriculum, then its gains should shrink substantially under this ordering. 
On the other hand, if our method preserves and reuses transferable skill neighborhoods, it should remain competitive even when the sequence begins with the hardest tasks.

\begin{table*}[ht]
    \centering
    \small
    \setlength{\tabcolsep}{5.2pt}
    \begin{tabular}{lp{1.60cm}cccccc}
        \toprule
        Method & Mean SR $\uparrow$ & TTT $\downarrow$ & BWT $\uparrow$ & Coverage $\uparrow$ & nBWT $\uparrow$ & TR $\uparrow$ \\
        \midrule
        \textsc{Scratch}
        & $0.359 \pm 0.03$   
        & $7.92 \pm 2.25$ 
        & $-0.291 \pm 0.024$ 
        & $0.35 \pm 0.05$ 
        & $-0.728 \pm 0.091$ 
        & $0.04 \pm 0.04$  \\

        \textsc{Scr.-Reuse} 
        & $0.514 \pm 0.04$   
        & $5.66 \pm 2.43$ 
        & $-0.055 \pm 0.007$ 
        & $0.31 \pm 0.05$ 
        & $\mathbf{-0.017 \pm 0.015}$ 
        & $\mathbf{0.31 \pm 0.06}$  \\

        \textsc{Finetune}
        & $0.590 \pm 0.10$   
        & $5.86 \pm 2.44$ 
        & $-0.379 \pm 0.037$ 
        & $0.62 \pm 0.05$ 
        & $-0.568 \pm 0.041$ 
        & $0.23 \pm 0.03$  \\

        \textsc{Finetune-Reset} 
        & $0.690 \pm 0.09$   
        & $4.56 \pm 2.33$ 
        & $-0.387 \pm 0.041$ 
        & $0.59 \pm 0.05$ 
        & $-0.613 \pm 0.052$ 
        & $0.21 \pm 0.02$  \\

        \textsc{EWC}
        & $0.364 \pm 0.08$   
        & $7.68 \pm 2.37$ 
        & $-0.169 \pm 0.052$ 
        & $0.42 \pm 0.09$ 
        & $-0.314 \pm 0.096$ 
        & $0.30 \pm 0.03$  \\

        \textsc{L2Init}
        & $0.242 \pm 0.04$   
        & $8.84 \pm 2.17$ 
        & $-0.172 \pm 0.046$ 
        & $0.19 \pm 0.06$ 
        & $-0.576 \pm 0.199$ 
        & $0.05 \pm 0.04$  \\
        
        \textsc{DFF} 
        & $0.398 \pm 0.03$  
        & $7.31 \pm 2.10$ 
        & $\mathbf{-0.020 \pm 0.009}$  
        & $0.06 \pm 0.04$ 
        & $-0.094 \pm 0.069$ 
        & $0.12 \pm 0.04$  \\

        \textsc{$S\&P$} 
        & $0.671 \pm 0.10$  
        & $4.79 \pm 2.34$  
        & $-0.395 \pm 0.005$ 
        & $0.62 \pm 0.09$ 
        & $-0.586 \pm 0.086$ 
        & $0.21 \pm 0.02$  \\

        \textsc{TeLAPA-Static} 
        & $0.890 \pm 0.04$   
        & $1.74 \pm 1.49$ 
        & $-0.493 \pm 0.068$ 
        & $0.63 \pm 0.04$ 
        & $-0.646 \pm 0.061$ 
        & $0.21 \pm 0.02$  \\

        \textsc{TeLAPA} 
        & $\mathbf{0.903 \pm 0.04}$   
        & $\mathbf{1.60 \pm 1.54}$ 
        & $-0.411 \pm 0.046$ 
        & $\mathbf{0.70 \pm 0.08}$ 
        & $-0.569 \pm 0.061$ 
        & $0.24 \pm 0.04$  \\

        \bottomrule
    \end{tabular}
    \caption{
    \textbf{Alternative curriculum 1: reverse anti-curriculum.}
    We report mean $\pm$ 95\% CI across 20 runs. We evaluate all single-model preservation baselines and TeLAPA variations following the Alternative Curriculum 1. 
    We use the same hyperparameter configuration for all methods as in the main paper results.
    }
    \label{tab:results_app_baselines_anti_curriculum}
    \vspace{-12pt}
\end{table*}

\paragraph{Alternative curriculum 2: skill-scrambled non-monotone order.}
The second alternative is a non-monotone, skill-scrambled sequence:
\[
C \rightarrow A \rightarrow E \rightarrow B \rightarrow D \rightarrow C' \rightarrow A' \rightarrow E' \rightarrow B' \rightarrow D'.
\]
Unlike the reverse ordering, this variant is not simply hardest-to-easiest. Instead, it deliberately breaks the smooth skill ladder by forcing abrupt shifts in task horizon and interaction structure. 
For example, the agent moves from key--door dependency to pure navigation, then to the longest-horizon maze, then back to short-horizon pickup, and finally to obstacle manipulation. 

\begin{table*}[ht]
    \centering
    \small
    \setlength{\tabcolsep}{5.2pt}
    \begin{tabular}{lp{1.60cm}cccccc}
        \toprule
        Method & Mean SR $\uparrow$ & TTT $\downarrow$ & BWT $\uparrow$ & Coverage $\uparrow$ & nBWT $\uparrow$ & TR $\uparrow$ \\
        \midrule
        \textsc{Scratch}
        & $0.356 \pm 0.03$   
        & $7.88 \pm 2.22$ 
        & $-0.377 \pm 0.046$ 
        & $0.35 \pm 0.08$ 
        & $-0.923 \pm 0.038$ 
        & $0.00 \pm 0.00$  \\

        \textsc{Scr.-Reuse} 
        & $0.513 \pm 0.03$   
        & $5.63 \pm 2.46$ 
        & $\mathbf{-0.055 \pm 0.008}$ 
        & $0.33 \pm 0.06$ 
        & $\mathbf{-0.011 \pm 0.012}$ 
        & $\mathbf{0.32 \pm 0.06}$  \\

        \textsc{Finetune}
        & $0.807 \pm 0.09$   
        & $3.77 \pm 2.00$ 
        & $-0.489 \pm 0.061$ 
        & $0.77 \pm 0.05$ 
        & $-0.775 \pm 0.058$ 
        & $0.18 \pm 0.08$  \\

        \textsc{Finetune-Reset} 
        & $0.726 \pm 0.08$   
        & $4.58 \pm 2.16$ 
        & $-0.557 \pm 0.063$ 
        & $0.75 \pm 0.04$ 
        & $-0.857 \pm 0.032$ 
        & $0.09 \pm 0.06$  \\

        \textsc{EWC}
        & $0.420 \pm 0.05$   
        & $6.44 \pm 2.54$ 
        & $-0.287 \pm 0.062$ 
        & $0.81 \pm 0.08$ 
        & $-0.345 \pm 0.064$ 
        & $0.22 \pm 0.07$  \\

        \textsc{L2Init}
        & $0.224 \pm 0.04$   
        & $8.98 \pm 2.16$ 
        & $-0.136 \pm 0.033$ 
        & $0.11 \pm 0.06$ 
        & $-0.479 \pm 0.187$ 
        & $0.01 \pm 0.02$  \\
        
        \textsc{DFF} 
        & $0.409 \pm 0.02$  
        & $8.36 \pm 2.03$ 
        & $-0.105 \pm 0.007$  
        & $0.15 \pm 0.04$ 
        & $-0.336 \pm 0.041$ 
        & $0.00 \pm 0.00$  \\

        \textsc{$S\&P$} 
        & $0.776 \pm 0.09$  
        & $3.92 \pm 2.01$  
        & $-0.550 \pm 0.076$ 
        & $0.81 \pm 0.02$ 
        & $-0.798 \pm 0.059$ 
        & $0.16 \pm 0.09$  \\

        \textsc{TeLAPA-Static} 
        & $0.915 \pm 0.04$   
        & $0.99 \pm 1.33$ 
        & $-0.688 \pm 0.027$ 
        & $\mathbf{0.92 \pm 0.05}$ 
        & $-0.789 \pm 0.046$ 
        & $0.15 \pm 0.05$  \\

        \textsc{TeLAPA} 
        & $\mathbf{0.927 \pm 0.04}$   
        & $\mathbf{0.95 \pm 1.30}$ 
        & $-0.671 \pm 0.035$ 
        & $0.91 \pm 0.04$ 
        & $-0.805 \pm 0.053$ 
        & $0.18 \pm 0.06$  \\
        \bottomrule
    \end{tabular}
    \caption{
    \textbf{Alternative curriculum 2: skill-scrambled non-monotone order.}
    We report mean $\pm$ 95\% CI across 20 runs. We evaluate all single-model preservation baselines and TeLAPA variations following the Alternative Curriculum 2. 
    We use the same hyperparameter configuration for all methods as in the main paper results.
    }
    \label{tab:results_app_baselines_scrambled_curriculum}
\end{table*}

This sequence tests whether TeLAPA requires gradual prerequisite accumulation, or whether it can still retain and exploit useful policies when tasks arrive in a less curriculum-like order.

\paragraph{Results.}
Task order still matters, but TeLAPA's advantage does not disappear under either alternative curriculum. 
Under the anti-curriculum, TeLAPA achieves the best Mean SR ($0.903$), fastest TTT ($1.60$), and highest Coverage ($0.70$), with TeLAPA-Static close behind. 
Under the skill-scrambled curriculum, the same pattern holds: TeLAPA again obtains the best Mean SR ($0.927$) and TTT ($0.95$), while TeLAPA-Static attains the highest Coverage ($0.92$) and TeLAPA remains nearly identical ($0.91$). 
Although some single-model baselines retain stronger BWT, nBWT, or TR values in specific cases, the overall picture is consistent across both reorderings: TeLAPA remains the strongest method on the main forward-performance metrics even when the original skill ladder is broken or reversed. 
This supports the claim that its gains are not tied to one favorable hand-designed curriculum, but to preserving and reusing transferable skill neighborhoods across task orders.


\subsection{Longer MiniGrid Curriculum Ablation}
\label{app:long_curriculum}

Task-order ablations in App.~\ref{app:curriculum_ablation} test whether \textsc{TeLAPA}'s advantage depends on the particular ordering of the original five-task sequence. 
However, a separate concern is that the original curriculum may be too short: with only five task families before revisits, the interference interval is limited, the archive library remains relatively small, and latent-space maintenance is tested across only a modest number of task boundaries. 
To address this, we introduce a longer MiniGrid curriculum that preserves the original skill-ladder sequence while adding three additional task families before revisits.

The long curriculum is:
\[
A \rightarrow B \rightarrow C \rightarrow D \rightarrow E \rightarrow F \rightarrow G \rightarrow H
\rightarrow A' \rightarrow B' \rightarrow C' \rightarrow D' \rightarrow E' \rightarrow F' \rightarrow G' \rightarrow H' .
\]
We increase the number of task boundaries from 10 to 16 and lengthen the interference gap before each revisit. 
As a result, the setting places additional pressure on both policy reuse and latent-space maintenance: a method must preserve useful archived policies over more intervening tasks, retrieve from a larger archive library, and remain robust to additional representation drift before revisiting earlier tasks.

\begin{table}[ht]
\centering
\small
\setlength{\tabcolsep}{6pt}
\begin{tabular}{clp{0.5\linewidth}}
\toprule
\textbf{Tag} & \textbf{Environment ID} & \textbf{Role in the long curriculum} \\
\midrule
F & \texttt{MiniGrid-RedBlueDoors-8x8-v0} 
& Adds ordered door interaction: the red door must be opened before the blue door. \\

G & \texttt{MiniGrid-KeyCorridorS3R3-v0} 
& Adds multi-room key search, locked-door access, and target-object pickup. \\

H & \texttt{MiniGrid-MultiRoom-N2-S4-v0} 
& Adds longer-horizon multi-room traversal through connected rooms and doors. \\
\bottomrule
\end{tabular}
\caption{\textbf{Additional MiniGrid Tasks.} 
    The original five-task sequence described in Table~\ref{tab:envs} are preserved and extended with three additional task families before revisits. 
    This increases both the number of task boundaries and the interference interval before each revisit.}
\label{tab:long_curriculum_tasks_envs}
\end{table}

The added environments were selected to introduce qualitatively different forms of interaction while remaining within the same MiniGrid observation and action interface used in the main experiments. 
\texttt{RedBlueDoors} adds order-sensitive door manipulation, 
\texttt{KeyCorridor} combines multi-room exploration with hidden-key retrieval and locked-door object pickup, and 
\texttt{MultiRoom} stresses longer-horizon navigation through connected rooms. 
Thus, the long curriculum is not merely a repetition of the original task ladder; it introduces additional sources of interference and retrieval ambiguity while preserving comparability with the main experimental setup.

\begin{table*}[ht]
    \centering
    \small
    \setlength{\tabcolsep}{5.0pt}
    \begin{tabular}{lcccccc}
        \toprule
        Method & Mean SR $\uparrow$ & TTT $\downarrow$ & BWT $\uparrow$ & Coverage $\uparrow$ & nBWT $\uparrow$ & TR $\uparrow$ \\
        \midrule
        \textsc{Scratch}
        & $0.417 \pm 0.01$   
        & $6.98 \pm 2.23$ 
        & $-0.344 \pm 0.017$ 
        & $0.47 \pm 0.04$ 
        & $-0.705 \pm 0.026$ 
        & $0.12 \pm 0.00$  \\
        
        \textsc{Scr.-Reuse} 
        & $0.573 \pm 0.01$   
        & $4.94 \pm 2.47$ 
        & $\mathbf{-0.048 \pm 0.004}$ 
        & $0.42 \pm 0.04$ 
        & $\mathbf{+0.002 \pm 0.013}$ 
        & $\mathbf{0.44 \pm 0.04}$  \\
        
        \textsc{Finetune}
        & $0.624 \pm 0.05$   
        & $5.01 \pm 2.20$ 
        & $-0.294 \pm 0.045$ 
        & $0.43 \pm 0.06$ 
        & $-0.592 \pm 0.045$ 
        & $0.14 \pm 0.02$  \\
        
        \textsc{Finetune-Reset} 
        & $0.504 \pm 0.06$   
        & $5.66 \pm 2.32$ 
        & $-0.315 \pm 0.057$ 
        & $0.46 \pm 0.07$ 
        & $-0.572 \pm 0.082$ 
        & $0.16 \pm 0.03$  \\
        
        \textsc{EWC}
        & $0.328 \pm 0.06$   
        & $7.19 \pm 2.54$ 
        & $-0.150 \pm 0.047$ 
        & $0.41 \pm 0.08$ 
        & $-0.202 \pm 0.084$ 
        & $0.26 \pm 0.07$  \\
        
        \textsc{L2Init}
        & $0.243 \pm 0.02$   
        & $8.99 \pm 2.20$ 
        & $-0.139 \pm 0.019$ 
        & $0.19 \pm 0.04$ 
        & $-0.235 \pm 0.116$ 
        & $0.12 \pm 0.00$  \\
        
        \textsc{DFF} 
        & $0.459 \pm 0.02$  
        & $7.27 \pm 2.32$ 
        & $-0.151 \pm 0.012$  
        & $0.21 \pm 0.04$ 
        & $-0.225 \pm 0.100$ 
        & $0.12 \pm 0.00$  \\
        
        \textsc{$S\&P$} 
        & $0.646 \pm 0.07$  
        & $4.59 \pm 2.10$  
        & $-0.337 \pm 0.037$ 
        & $0.45 \pm 0.04$ 
        & $-0.656 \pm 0.054$ 
        & $0.12 \pm 0.00$  \\
        
        \textsc{TeLAPA-Static} 
        & $0.573 \pm 0.03$   
        & $5.53 \pm 3.00$ 
        & $-0.287 \pm 0.033$ 
        & $0.44 \pm 0.05$ 
        & $-0.563 \pm 0.073$ 
        & $0.16 \pm 0.03$  \\

        \textsc{TeLAPA} 
        & $\mathbf{0.687 \pm 0.06}$   
        & $\mathbf{4.26 \pm 2.57}$ 
        & $-0.372 \pm 0.054$ 
        & $\mathbf{0.49 \pm 0.07}$ 
        & $-0.581 \pm 0.062$ 
        & $0.15 \pm 0.03$  \\
        \bottomrule
    \end{tabular}
    \caption{\textbf{Long-curriculum results.}
    We report mean $\pm$ 95\% CI across 20 runs. 
    We evaluate all single-model preservation baselines and TeLAPA variation.
    We use the same hyperparameter configuration for all methods as in the main paper results.}
    \label{tab:long_curriculum_results}
    \vspace{-12pt}
\end{table*}

\vspace{-1.0em}
\paragraph{Results.}
The long-curriculum results show that \textsc{TeLAPA}'s advantage does not disappear when the task stream is extended. 
Across the 16-task sequence, \textsc{TeLAPA} achieves the best Mean SR ($0.687 \pm 0.06$), the fastest TTT ($4.26 \pm 2.57$), and the highest Coverage ($0.49 \pm 0.07$). 
This indicates that preserving and selecting among reusable policy neighborhoods remains beneficial even when the archive library is larger, the number of task boundaries increases, and revisits occur after a longer interference interval. 
The strongest single-model competitor is \textsc{S\&P}, which obtains competitive Mean SR ($0.646 \pm 0.07$) and TTT ($4.59 \pm 2.10$), but it does not match \textsc{TeLAPA}'s overall recoverability profile and has substantially lower threshold retention. 
\textsc{Scratch-Reuse} remains strongest on BWT, nBWT, and TR, which is expected because it stores one task-specific policy and reuses it directly on revisits. 
However, it achieves lower Mean SR, slower TTT, and lower Coverage than \textsc{TeLAPA}, suggesting that retaining a single policy per task is less effective than maintaining a broader archive of transfer candidates when the curriculum becomes longer. 
Finally, \textsc{TeLAPA} outperforms \textsc{TeLAPA-Static} on Mean SR, TTT, and Coverage, supporting the importance of online latent-space maintenance rather than archive storage alone. 
Overall, these results directly address the concern that the original A--E curriculum may be too short or unusually favorable.
\textsc{TeLAPA} continues to perform best on the metrics most directly tied to long-horizon recoverability under a longer and more interference-heavy task stream.

\subsection{Episodic Intrinsic Reward Ablation}
\label{app:intrinsic_reward_ablation}
Intrinsic rewards are a standard tool for sparse-reward exploration, but their effect in our setting is not obvious. 
In TeLAPA, the relevant question is not whether extra diversity can be produced, but whether the induced diversity remains \emph{skill-aligned} with later transfer targets. 
This distinction matters because an exploration bonus may help an agent discover additional competent behaviors on the current task, while simultaneously shifting the stored neighborhood toward exploratory modes that are less reusable on later revisits.
Therefore, we study an intrinsic-reward ablation to test whether adding exploration pressure during policy optimization is uniformly beneficial for continual transfer, or whether it changes the archive in ways that are only conditionally useful. 

We base our choice of an \emph{episodic} count bonus on a series of recent works.
First, \citet{wang2023revisiting} revisits intrinsic rewards in procedurally generated environments and compare combinations of several \emph{lifelong} bonuses, including ICM \citep{pathak2017curiosity}, RIDE \citep{raileanu2020ride}, RND \citep{burda2018exploration}, and BeBold \citep{zhang2020bebold}, with simple \emph{episodic} bonuses such as episodic count and episodic visitation. 
Second, \citet{henaff2023study} analyzes the tradeoff between global and episodic bonuses in contextual MDPs, showing that episodic bonuses are especially effective when there is limited shared structure across episodes.

\paragraph{Formulation.}
We use an \emph{episodic count-based exploration bonus}. 
Let $\phi(s_t)$ denote the hashed state representation used for counting, and let $N_{\mathrm{ep}}(\phi(s_t))$ be the number of visits to that state \emph{within the current episode}.
\begin{equation}
\label{eq:intrinsic_reward}
r^{\mathrm{tot}}_t
=
r^{\mathrm{ext}}_t
+
r^{\mathrm{int}}_t,
\qquad
r^{\mathrm{int}}_t
=
\beta_{\mathrm{int}}
\big/
\sqrt{N_{\mathrm{ep}}(\phi(s_t))}.
\end{equation}
In our implementation, we augment the environment reward during \emph{during PPO training only} and remove at the time of evaluation. 
The episodic count table is reset at every environment reset. 

\begin{table*}[ht]
    \centering
    \small
    \setlength{\tabcolsep}{5.0pt}
    \begin{tabular}{lcccccc}
        \toprule
        Method & Mean SR $\uparrow$ & TTT $\downarrow$ & BWT $\uparrow$ & Coverage $\uparrow$ & nBWT $\uparrow$ & TR $\uparrow$ \\
        \midrule
        \textsc{Scratch}
        & $0.302 \pm 0.00$   
        & $7.74 \pm 2.34$ 
        & $-0.303 \pm 0.009$ 
        & $0.24 \pm 0.04$ 
        & $-0.911 \pm 0.051$ 
        & $0.00 \pm 0.00$  \\
        
        \textsc{Scr.-Reuse} 
        & $0.357 \pm 0.02$   
        & $7.26 \pm 2.52$ 
        & $-0.079 \pm 0.005$ 
        & $0.25 \pm 0.04$ 
        & $\mathbf{-0.015 \pm 0.013}$ 
        & $0.25 \pm 0.04$  \\
        
        \textsc{Finetune}
        & $0.650 \pm 0.10$   
        & $4.95 \pm 2.35$ 
        & $-0.462 \pm 0.041$ 
        & $0.72 \pm 0.06$ 
        & $-0.582 \pm 0.068$ 
        & $0.25 \pm 0.05$  \\
        
        \textsc{Finetune-Reset} 
        & $0.698 \pm 0.09$   
        & $4.50 \pm 2.32$ 
        & $-0.460 \pm 0.051$ 
        & $0.71 \pm 0.05$ 
        & $-0.569 \pm 0.071$ 
        & $0.23 \pm 0.05$  \\
        
        \textsc{EWC}
        & $0.233 \pm 0.05$   
        & $8.75 \pm 2.25$ 
        & $-0.156 \pm 0.022$ 
        & $0.26 \pm 0.08$ 
        & $-0.351 \pm 0.146$ 
        & $0.20 \pm 0.02$  \\
        
        \textsc{L2Init}
        & $0.303 \pm 0.05$   
        & $8.23 \pm 2.07$ 
        & $-0.303 \pm 0.012$ 
        & $0.28 \pm 0.04$ 
        & $-0.854 \pm 0.019$ 
        & $0.01 \pm 0.03$  \\
        
        \textsc{DFF} 
        & $0.710 \pm 0.02$  
        & $4.20 \pm 2.24$ 
        & $\mathbf{-0.059 \pm 0.035}$  
        & $0.46 \pm 0.06$ 
        & $-0.414 \pm 0.065$ 
        & $\mathbf{0.37 \pm 0.06}$  \\
        
        \textsc{$S\&P$} 
        & $\mathbf{0.724 \pm 0.06}$  
        & $4.25 \pm 2.21$  
        & $-0.426 \pm 0.027$ 
        & $\mathbf{0.73 \pm 0.06}$ 
        & $-0.529 \pm 0.045$ 
        & $0.26 \pm 0.01$  \\
        
        \textsc{TeLAPA-Static} 
        & $0.668 \pm 0.09$   
        & $3.90 \pm 2.32$ 
        & $-0.178 \pm 0.086$ 
        & $0.40 \pm 0.06$ 
        & $-0.593 \pm 0.104$ 
        & $0.35 \pm 0.11$  \\

        \textsc{TeLAPA} 
        & $0.681 \pm 0.08$   
        & $\mathbf{3.79 \pm 2.43}$ 
        & $-0.197 \pm 0.069$ 
        & $0.32 \pm 0.07$ 
        & $-0.684 \pm 0.128$ 
        & $0.26 \pm 0.06$  \\
        \bottomrule
    \end{tabular}
    \caption{
    \textbf{Episodic Intrinsic Reward Ablation.}
    We report mean $\pm$ 95\% CI across 20 runs. We introduce Episodic Intrinsic Reward in all single-model preservation baselines and TeLAPA variations following Eq.~\ref{eq:intrinsic_reward}. We use the same hyperparameter configuration for all methods as in the main paper results.
    }
    \label{tab:results_app_baselines_intrinsic}
\end{table*}

The intrinsic-reward ablation shows that additional exploration pressure is not uniformly helpful for continual transfer. 
Although several methods improve under this setting, the best results are obtained by $S\&P$ and DFF rather than by TeLAPA variants. 
TeLAPA and TeLAPA-Static still adapt relatively quickly, but they do not achieve the strongest Mean SR, Coverage, or TR, indicating that the extra exploratory behaviors induced by the bonus are not necessarily the most reusable ones. 
Overall, these results support the view that diversity alone is insufficient: what matters is whether exploration produces \emph{skill-aligned} neighborhoods that remain useful for later retrieval and transfer, since additional exploratory pressure can broaden the archive in ways that reduce later reuse when the induced diversity is not sufficiently skill-aligned.

\subsection{Cross-Archive Elite Injection Ablation}
\label{app:elite_injection_ablation}

A central claim of the paper is that useful diversity is not arbitrary diversity, but \emph{skill-aligned} diversity. 
MAP-Elites already expands local neighborhoods through mutation and selection, but we can additionally force the search to occasionally branch from elites originating outside the current task archive. 
Therefore, we study an archive-level ablation that explicitly perturbs how diversity is generated during illumination to test whether occasionally importing parents from outside the current archive improves useful coverage, or instead introduces behaviorally diverse but weakly reusable modes.

\paragraph{Injection mechanism.}
During illumination, let $p_{\mathrm{inj}}$ denote the elite-injection probability. To avoid indiscriminate perturbations, at each parent-selection step, with probability $p_{\mathrm{inj}}$ we replace the standard parent choice with a parent drawn from a curated pool of elites collected from other archives. When $p_{\mathrm{inj}}=0$, the method reduces to the default no-injection setting.
\begin{equation}
\label{eq:injection_elite}
\mathrm{score}(z_i)
=
d_{\min}(z_i,\mathcal{A}_{\mathrm{curr}})
+
\lambda \,\tilde f_i,
\end{equation}
where $d_{\min}(z_i,\mathcal{A}_{\mathrm{curr}})$ is the minimum latent-space distance from candidate $i$ to the current archive, and $\tilde f_i$ is the candidate's normalized source fitness within the sampled subset. 
Thus, injected parents are chosen to occupy regions that are under-covered by the current archive while avoiding extremely weak candidates.

\begin{table*}[ht]
    \centering
    \small
    \setlength{\tabcolsep}{5.0pt}
    \begin{tabular}{lp{1.65cm}cccccc}
        \toprule
        Method & Mean SR $\uparrow$ & TTT $\downarrow$ & BWT $\uparrow$ & Coverage $\uparrow$ & nBWT $\uparrow$ & TR $\uparrow$ \\
        \midrule
     
        \textsc{$p_{\mathrm{inj}}=0$} 
        & $\mathbf{0.706 \pm 0.08}$  
        & $\mathbf{3.35 \pm 2.09}$  
        & $-0.303 \pm 0.086$ 
        & $0.50 \pm 0.09$ 
        & $-0.629 \pm 0.071$ 
        & $0.25\pm 0.05$  \\

        \textsc{$p_{\mathrm{inj}}=0.1$} 
        & $0.636 \pm 0.09$   
        & $4.16 \pm 2.37$ 
        & $-0.328 \pm 0.064$ 
        & $\mathbf{0.53 \pm 0.08}$ 
        & $-0.637 \pm 0.085$ 
        & $0.22 \pm 0.05$  \\

        \textsc{$p_{\mathrm{inj}}=0.2$} 
        & $0.638 \pm 0.09$   
        & $4.06 \pm 2.38$ 
        & $-0.327 \pm 0.065$ 
        & $0.52 \pm 0.09$ 
        & $-0.707 \pm 0.062$ 
        & $0.20 \pm 0.04$  \\

        \textsc{$p_{\mathrm{inj}}=0.3$} 
        & $0.598 \pm 0.09$   
        & $4.84 \pm 2.37$ 
        & $-0.256 \pm 0.076$ 
        & $0.42 \pm 0.09$ 
        & $\mathbf{-0.597 \pm 0.074}$ 
        & $0.25 \pm 0.05$  \\

        \textsc{Intr.+$p_{\mathrm{inj}}=0.2$} 
        & $0.673 \pm 0.09$   
        & $4.25 \pm 2.50$ 
        & $\mathbf{-0.175 \pm 0.067}$ 
        & $0.29 \pm 0.07$ 
        & $-0.717 \pm 0.145$ 
        & $\mathbf{0.26 \pm 0.08}$  \\
        \bottomrule
    \end{tabular}
    \caption{
    \textbf{TeLAPA Cross-Archive Elite Injection Ablation.}
    We report mean $\pm$ 95\% CI across 20 runs. We introduce the Elite Injection mechanism with various elite-injection probabilities $p_{\mathrm{inj}} \in [0.1, 0.2, 0.3]$ following Eq.~\ref{eq:injection_elite}. 
    We use the same hyperparameter configuration for all methods as in the main paper results.
    Additionally we combine the best $p_{\mathrm{inj}}$ value with Episodic Intrinsic Reward (Intr.) to evaluate the impact of both ablation studies used together.
    $p_{\mathrm{inj}}=0$ indicates main paper baseline as seen in Tab.~\ref{tab:results_main_baselines}.
    }
    \label{tab:results_app_baselines_injection}
    \vspace{-10pt}
\end{table*}

We interpret elite injection as an \emph{archive-shaping intervention} to isolate whether \emph{externally introduced} diversity helps or creates noise. 
Introducing it at $p_{\mathrm{inj}}\in\{0.1,0.2,0.3\}$ does not improve overall performance: while small injection slightly increases coverage (best at $p_{\mathrm{inj}}=0.1$ with $0.53$), it consistently lowers Mean SR, slows adaptation, and generally reduces transfer ratio. 
At higher injection rates, these degradations become more pronounced, suggesting that imported parents are behaviorally diverse but not sufficiently aligned with the target task's skill structure. 
Thus, proving that diversity must remain structured by the target-relevant geometry of the archive in order to help long-horizon revisit recovery.
Combining injection with intrinsic reward partially improves BWT and TR, but still fails to recover the baseline's overall balance and yields substantially lower coverage. 
\section{Geometry diagnostics in the shared behavior space}
\vspace{-0.5em}
\label{app:geometry_diagnostics}

We expand the analysis in Section~\ref{subsec:stepping_stones} by testing whether per-task archives occupy structured and transfer-relevant regions of the shared latent behavior space through a geometry perspective.
Within each run, all task archives are embedded in a common standardized latent space, which enables cross-task comparisons of archive location, overlap, and nearest cross-task attachment.\footnote{Because the embedder and normalization pipeline are non-stationary and run-dependent, raw latent coordinates are not assumed aligned across seeds. Therefore, we aggregate \emph{scalar} geometry statistics across runs (means and confidence intervals), and we pool only those quantities that can be made dimensionless.} 
For each task archive $\mathcal{A}_k$, we compute the archive centroid ($c_k$), whose distances ($D_{k\ell}$) provide a measure of inter-task separation:
\begin{equation}
c_k = \frac{1}{N_k}\sum_{i=1}^{N_k}\tilde{z}_{k,i},
\qquad
D_{k\ell} = \lVert c_k - c_\ell \rVert_2  ;
\end{equation}
However, centroid distances alone ignore the spread within the task. We therefore compute a within-task dispersion (RMS radius) and define a \emph{separation ratio} that normalizes distance by the spread within the task:
\begin{equation}
R_k \;=\; \sqrt{\frac{1}{N_k}\sum_{i=1}^{N_k}\lVert \tilde{z}_{k,i} - c_k \rVert_2^2};
\qquad
S_{k\ell} \;=\; \frac{\lVert c_k - c_\ell\rVert_2}{\sqrt{R_k^2 + R_\ell^2} + \epsilon}.
\label{eq:separation_ratio}
\end{equation}

\begin{figure}[ht]
  \centering
  \includegraphics[width=\linewidth]{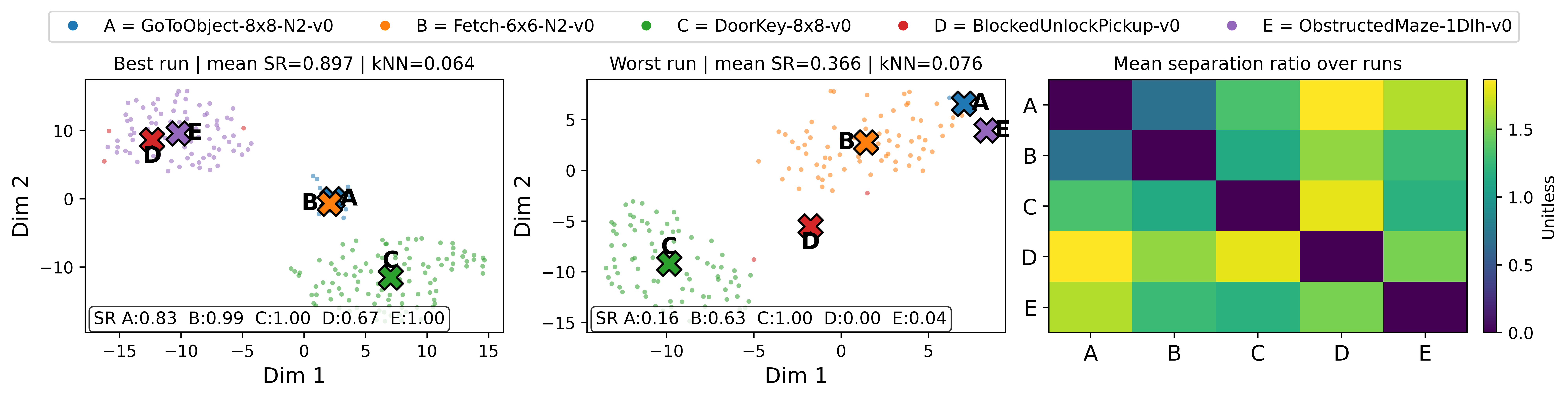}
  \caption{
  \textbf{Cross-task geometry in the shared latent space.}
  \textbf{Left:} t-SNE of elites from the best run (highest mean SR across tasks).
  \textbf{Middle:} same visualization for the worst run (lowest mean SR across tasks).
  \textbf{Right:} mean separation ratio matrix $S_{k\ell}$ (Eq.~\ref{eq:separation_ratio}), measuring centroid distance normalized by within-task dispersion.}  
  \label{fig:v2_joint_tsne_heatmap}
\end{figure}

Large $S_{k\ell}$ indicates that two task archives are well separated relative to their within-task spread. We also visualize two representative runs in the shared latent behavior space: the run with the highest mean success rate (SR) across tasks and the run with the lowest mean SR. 
This figure provides a meaningful organized look into the latent space, allowing us to analyze whether two tasks in nearby regions are more likely to transfer to the other.
The first two panels of Fig.~\ref{fig:v2_joint_tsne_heatmap} show 2D t-SNE projections of archived elites\footnote{t-SNE is nonlinear and stochastic: its axes are arbitrary and global distances in 2D should not be over-interpreted.}, colored by task, with task centroids marked by ``X''.

Figure~\ref{fig:v2_joint_tsne_heatmap} indicates a structured shared space and suggests that high performance runs are less associated with maximal task separation than with coherent local organization.
In the best run, archives form compact task-specific neighborhoods with interpretable adjacencies: A and B lie very close, C occupies a distinct region, while E and D sit far from the rest but close together.
In the worst run, the structure is less favorable: B spreads broadly across the space, while several low-performing tasks occupy smaller or poorly aligned regions, even though some centroid distances become larger.
Thus, larger global separation alone is not a sufficient indicator of useful geometry.
What appears more important is whether tasks form compact, recoverable neighborhoods, and meaningful local adjacencies that can support retrieval.

\begin{table}[ht]
\centering
\small
\setlength{\tabcolsep}{5.5pt}
\begin{tabular}{lccccc}
\toprule
\textbf{Method} & $A'$ & $B'$ & $C'$ & $D'$ & $E'$ \\
\midrule
\textsc{Scratch}        & 0.258 & 0.480 & 1.00 & 0.00 & 0.129 \\
\textsc{Scr.-Reuse}     & 0.156 & 0.497 & 1.00 & 0.00 & 0.97 \\
\textsc{Finetune}       & 0.601 & 0.822 & 0.80 & 0.168 & 0.55 \\
\textsc{Finetune-Reset} & 0.628 & 0.846 & 0.752 & 0.20 & 0.454 \\
\textsc{EWC}            & 0.113 & 0.472 & 0.11 & 0.00 & 0.008 \\
\textsc{L2Init}         & 0.091 & 0.478 & 0.027 & 0.003 & 0.001 \\
\textsc{DFF}            & 0.289 & 0.65 & 0.995 & 0.003 & 0.003 \\
\textsc{S\&P}           & \textbf{0.737} & \textbf{0.928} & 0.451 & 0.00 & 0.20 \\
\textsc{TeLAPA-Static}  & 0.277 & 0.592 & 1.00 & 0.152 & 0.513 \\
\textsc{TeLAPA}         & 0.526 & 0.698 & $\textbf{1.00}^*$ & \textbf{0.357} & \textbf{0.99} \\
\bottomrule
\end{tabular}
\caption{\textbf{Revisit Mean SR (Continual Learning signal).}
Per-task mean Success Rate on revisited tasks after the archive has been illuminated across previous tasks. 
$^*$ indicates that multiple methods achieve 100\% success rate on that task.}
\label{tab:revisit_main}
\vspace{-10pt}
\end{table}

The latent-space visualization provided by Fig.~\ref{fig:v2_joint_tsne_heatmap} suggested that useful transfer is associated less with maximal global separation than with compact task-specific neighborhoods and meaningful local adjacencies. 
Table~\ref{tab:revisit_main} provides a per-task functional look on the performance exclusive to the revisit results originally presented in Table.~\ref{tab:results_main_baselines}. 
\textsc{TeLAPA} is the strongest or tied strongest method on the harder revisits, reaching the best performance on $D'$ and $E'$ and matching the ceiling on $C'$, while also remaining competitive on $A'$ and $B'$. 
By contrast, methods that retain only a single solution or rely primarily on parameter regularization show much less consistent revisit behavior, often collapsing on some revisited tasks despite sometimes performing well on others. 
A particularly informative comparison is with \textsc{TeLAPA-Static}: although a fixed archive already helps on several revisits, the full online-maintained variant is substantially better on $D'$ and $E'$, which suggests that preserving diverse policies is not by itself sufficient; keeping those policies geometrically aligned in the updated shared space also matters. 
Likewise, \textsc{S\&P} attains the best scores on $A'$ and $B'$ but fails to generalize across the full revisit set, indicating that strong local recovery on a subset of tasks does not imply a reusable cross-task organization.

\section{Compute and Memory Overhead}
\label{app:compute_memory_overhead}

\textsc{TeLAPA} preserves a structured archive of policy neighborhoods rather than a single evolving policy. 
This additional structure introduces extra computation and storage relative to single-model baselines. 
Therefore, we measure the compute and memory footprint of each method by measuring a small set of metrics that separate training budget, method-specific overhead, runtime memory, and persistent storage. 
All measurements are collected on CPU-only runs using the same training configuration and curriculum as the main experiments.
 
\paragraph{Extra method environment steps.} 
We count only environment interactions introduced by the method outside the common PPO budget. 
For example, for \textsc{EWC} this includes Fisher-estimation rollouts, while for \textsc{TeLAPA} this includes archive illumination and few-shot transfer probing. 
Routine evaluation used for logging and metric computation is not counted as method-specific overhead because it is applied uniformly across methods.

\paragraph{Wall-clock time.}
We measure the total elapsed runtime of a full seed, from the start of the task sequence to final saving of results. 
We also report the wall-clock time relative to \textsc{Finetune}:
\[
\mathrm{RelTime}(m)
=
\frac{\mathrm{WallClock}(m)}
     {\mathrm{WallClock}(\textsc{Finetune})}.
\]

\paragraph{Peak CPU RAM.} 
This is the maximum resident set size observed during the run, including child processes created by vectorized environments. 
This measures whether a method requires substantially more runtime memory to fit on the same hardware. 

\paragraph{Disk storage.} 
This is the persistent storage footprint of the run directory after completion. 
This captures saved models, logs, archives, and method-specific artifacts. 

\paragraph{Stored method state} 
We isolate the persistent algorithmic state beyond the active PPO model and generic logs. 
For single-model methods this quantity is small or zero; for \textsc{Scratch-Reuse} it corresponds to the stored task policy bank; for \textsc{L2Init} it corresponds to the initialization anchor; for \textsc{EWC} it corresponds to the Fisher and parameter snapshots; and for \textsc{TeLAPA} it corresponds to the policy archive, trajectory banks, latent descriptors, embedder state, and archive metadata.

\begin{table*}[ht]
\centering
\small
\setlength{\tabcolsep}{4pt}
\begin{tabular}{lccccc}
\toprule
\textbf{Method} 
& \textbf{Extra method} 
& \textbf{Wall-clock} 
& \textbf{Rel. time} 
& \textbf{Peak CPU} 
& \textbf{Stored state} \\
& \textbf{env. steps} 
& \textbf{time} 
& \textbf{vs. FT} 
& \textbf{RAM} 
& \textbf{size} \\
\midrule
\textsc{Scratch}        & $0.00$M & $13.81 \pm 1.44$h & $0.90 \pm 0.09\times$ & $19.75 \pm 1.65$GB & $0.0$MB \\
\textsc{Scratch-Reuse}  & $0.00$M & $13.04 \pm 2.01$h & $0.84 \pm 0.13\times$ & $10.27 \pm 0.05$GB & $3.2$MB \\
\textsc{Finetune}       & $0.00$M & $15.43 \pm 0.30$h & $1.00\times$ & $10.02 \pm 0.03$GB & $0.0$MB \\
\textsc{Finetune-Reset} & $0.00$M & $15.20 \pm 0.27$h & $0.99 \pm 0.02\times$ & $10.26 \pm 0.04$GB & $0.0$MB \\
\textsc{EWC}            & $0.02$M & $15.34 \pm 0.20$h & $0.99 \pm 0.02\times$ & $10.22 \pm 0.03$GB & $1.3$MB \\
\textsc{L2Init}         & $0.00$M & $16.42 \pm 0.19$h & $1.07 \pm 0.03\times$ & $10.33 \pm 0.03$GB & $0.6$MB \\
\textsc{DFF}            & $0.00$M & $15.02 \pm 0.11$h & $0.97 \pm 0.02\times$ & $10.02 \pm 0.03$GB & $0.0$MB \\
\textsc{S\&P}           & $0.00$M & $15.31 \pm 0.21$h & $0.99 \pm 0.02\times$ & $14.79 \pm 0.05$GB & $0.0$MB \\
\textsc{TeLAPA-Static}  & $116.34 \pm 5.23$M & $18.77 \pm 1.66$h & $1.22 \pm 0.11\times$ & $52.02 \pm 0.71$GB & $188.6 \pm 43.9$MB \\
\textsc{TeLAPA}         & $86.29 \pm 4.77$M & $20.07 \pm 1.51$h & $1.30 \pm 0.10\times$ & $53.33 \pm 1.54$GB & $140.4 \pm 27.3$MB \\
\bottomrule
\end{tabular}
\caption{
\textbf{Compute and memory overhead across methods.}
Extra method environment steps count only method-specific interactions outside that common budget, excluding routine evaluation used for logging and metric computation. 
Wall-clock time is measured for the full run, and relative time is normalized by \textsc{Finetune}. 
Peak CPU RAM reports maximum resident memory during the run. 
Stored state size reports algorithm-specific persistent state beyond the active PPO model and generic logs. 
Values are reported as mean $\pm$ 95\% confidence interval across 20 runs.
}
\label{tab:compute_memory_overhead}
\end{table*}

\paragraph{\textsc{TeLAPA}'s additional runtime.}
We also record three \textsc{TeLAPA}-specific timers: 
\begin{itemize}
    \item \textbf{Archive illumination time:} measures the post-training MAP-Elites phase used to populate the task archive.
    \item \textbf{Few-shot origin-selection time:} measures transfer-time candidate evaluation and short adaptation probes.
    \item \textbf{Embedder-maintenance time:} measures anchor/replay updates, normalization, and archive re-embedding.    
\end{itemize}
These quantities are not separate methods, but a decomposition of the total \textsc{TeLAPA} wall-clock time.

\begin{table}[ht]
\centering
\small
\begin{tabular}{lc}
\toprule
\textbf{\textsc{TeLAPA} component} & \textbf{Wall-clock time} \\
\midrule
Archive illumination       & $2.99 \pm 0.33$h \\
Few-shot origin selection  & $2.23 \pm 0.21$h \\
Embedder maintenance       & $2.57 \pm 0.23$h \\
Other training/evaluation  & $12.28 \pm 0.95$h \\
\bottomrule
\end{tabular}
\caption{
    \textbf{Breakdown of \textsc{TeLAPA}'s runtime overhead.} 
    The illumination phase is expected to be the dominant additional cost because it evaluates perturbed policies to construct the per-task archive. 
    Few-shot selection and embedder maintenance measure transfer-time retrieval and latent-space maintenance costs, respectively.
}
\label{tab:telapa_runtime_breakdown}
\end{table}

\paragraph{Interpretation.}
Table~\ref{tab:compute_memory_overhead} shows that \textsc{TeLAPA}'s improved recoverability comes with a clear but quantifiable overhead. 
By construction, all methods use the same main PPO training budget. 
The extra environment-step column isolates method-specific interactions outside the shared training budget. 
Single-model baselines introduce essentially no extra method-specific environment interactions, except for \textsc{EWC}'s small Fisher-estimation pass ($0.02$M steps). 
In contrast, \textsc{TeLAPA-Static} and \textsc{TeLAPA} add $116.34 \pm 5.23$M and $86.29 \pm 4.77$M method-specific steps, respectively, reflecting the cost of archive illumination, transfer-time probing, and, for full \textsc{TeLAPA}, latent-space maintenance. 
Interestingly, \textsc{TeLAPA-Static} incurs more extra environment interactions than \textsc{TeLAPA}. 
This does not imply that \textsc{TeLAPA-Static} performs more algorithmic work by design. Instead, those extra environment steps count realized rollout lengths where unsuccessful MiniGrid policies often run until timeout, and weaker archive candidates can require more environment transitions per fixed evaluation episode. 
Therefore, the higher extra-step count for \textsc{TeLAPA-Static} is consistent with less reliable policies producing longer evaluation rollouts, while full \textsc{TeLAPA}'s latent-space maintenance adds compute time but can reduce the realized rollout cost of archive and transfer evaluations.

On the other hand, the increase in method-specific environment interactions translates into a more moderate wall-clock increase.
\textsc{TeLAPA} takes $20.07 \pm 1.51$h, or $1.30 \pm 0.10\times$ the \textsc{Finetune} runtime, while \textsc{TeLAPA-Static} takes $18.77 \pm 1.66$h, or $1.22 \pm 0.11\times$ \textsc{Finetune}. 
Thus, the main runtime cost is measurable but not proportional to the raw increase in extra environment interactions.

The larger practical cost is peak CPU memory. 
Most single-model baselines remain near $10$GB peak RAM, whereas \textsc{TeLAPA-Static} and \textsc{TeLAPA} peak at $52.02 \pm 0.71$GB and $53.33 \pm 1.54$GB, respectively. 
This indicates that archive-based preservation requires substantially more runtime memory than preserving a single active policy. 
At the same time, persistent method-specific storage remains modest in absolute terms. \textsc{TeLAPA} stores $140.4 \pm 27.3$MB and \textsc{TeLAPA-Static} stores $188.6 \pm 43.9$MB of additional method state, compared to at most a few MB for the strongest storage-based baselines. 
Therefore, while not being a dominant bottleneck, \textsc{TeLAPA}'s storage overhead is non-negligible.

Table~\ref{tab:telapa_runtime_breakdown} further decomposes the full \textsc{TeLAPA} runtime. 
Archive illumination is the largest individual \textsc{TeLAPA}-specific component at $2.99 \pm 0.33$h, but few-shot origin selection ($2.23 \pm 0.21$h) and latent-space embedder maintenance ($2.57 \pm 0.23$h) are also substantial. 
This suggests that the overhead is distributed across the three operations that distinguish \textsc{TeLAPA} from single-policy baselines: 
(i) constructing policy neighborhoods, 
(ii) selecting among prior archive elites, and 
(iii) maintaining a reusable latent-space coordinate system. 
Overall, \textsc{TeLAPA} improves success rate, time-to-threshold, coverage, and threshold retention, but does so by spending additional interactions and using substantially more runtime memory than single-model methods.
\section{Continuous-Control Dynamics-Shift Evaluation in MuJoCo}
\label{app:mujoco_friction}

To evaluate \textsc{TeLAPA}'s archive-based transfer mechanism in continuous-control domains, we add a MuJoCo dynamics-shift experiment. 
MuJoCo is a physics simulator commonly used for robotic locomotion and control benchmarks \citep{todorov2012mujoco}, and the Gymnasium MuJoCo suite provides standardized continuous-control environments such as \texttt{HalfCheetah}, \texttt{Hopper}, and \texttt{Walker2d} \citep{towers2024gymnasium}. 
Unlike the MiniGrid experiments, these tasks have continuous observation and action spaces, dense rewards, and locomotion dynamics governed by contact-rich physical simulation. 
Therefore, this setting provides a complementary stress test where we keep the morphology fixed within each experiment and induce non-stationarity by changing the contact dynamics.

\paragraph{Dynamics-shift curriculum.}
For each MuJoCo environment, we construct a cyclic curriculum by varying the ground-contact friction scale \citep{dohare2024loss}. 
Let $e \in \{\texttt{HalfCheetah-v5}, \texttt{Hopper-v5}, \texttt{Walker2d-v5}\}$ denote the base environment. 
For each $e$, we define five friction regimes,
\[
    \mathcal{F} = \{0.50,\;0.75,\;1.00,\;1.25,\;1.50\},
\]
where the scalar value rescales the sliding-friction coefficient of the agent. 
Then, we train on the cyclic sequence
\[
    F_{0.50} \rightarrow F_{0.75} \rightarrow F_{1.00} \rightarrow F_{1.25} \rightarrow F_{1.50}
    \rightarrow
    F'_{0.50} \rightarrow F'_{0.75} \rightarrow F'_{1.00} \rightarrow F'_{1.25} \rightarrow F'_{1.50}.
\]
This gives two passes through five dynamics regimes, yielding ten task visits per MuJoCo environment. 
The first pass measures adaptation to new physical dynamics, while the second pass measures recovery under revisits, analogous to the revisit structure used in the MiniGrid curriculum.

\paragraph{Why friction shifts preserve the \textsc{TeLAPA} interface.}
We run each MuJoCo environment independently. 
We do not transfer a policy from \texttt{HalfCheetah} to \texttt{Hopper} or to \texttt{Walker2d}. 
Instead, each experiment keeps the same morphology, observation dimension, action dimension, and PPO policy architecture, while varying only the friction regime. 
This design is important because \textsc{TeLAPA} stores full policy parameter vectors in its archives. 
Changing between different MuJoCo morphologies would generally change the observation and action dimensions, making archived policies shape-incompatible. 
By contrast, same-morphology friction shifts preserve policy compatibility and directly test whether archived policy neighborhoods remain useful under continuous-control dynamics changes.

\paragraph{Trajectory descriptors.}
The MiniGrid version of \textsc{TeLAPA} embeds episode trajectories using low-dimensional behavioral traces derived from position, action, and event features. 
For MuJoCo, we replace these symbolic/event features with a continuous-control trajectory sketch while keeping the same trajectory interface used by the latent embedder. 
At each timestep, we compress the MuJoCo transition into an $24$-dimensional feature vector containing summary statistics of the observation, summary statistics of the action, the instantaneous reward, the cumulative episode return, and normalized time. 
The resulting trajectory set has the same format as in the MiniGrid implementation, i.e.,
\[
    \tau = \{x_t\}_{t=1}^{T}, 
    \qquad x_t \in \mathbb{R}^{24},
\]
which allows the same trajectory embedder, archive re-embedding code, anchor/replay stores, and latent-space maintenance procedure to be used without changing the core \textsc{TeLAPA} algorithm.

\paragraph{Return-based competence.}
MiniGrid provides a natural binary success signal, so archive gates and transfer metrics can use success rate directly. 
MuJoCo tasks instead provide dense returns. 
To reuse the same archive and selection logic, we define a normalized return competence score,
\[
    \widetilde{\mathrm{SR}}_e(\pi)
    =
    \mathrm{clip}\left(
    \frac{\mathbb{E}[R_e(\pi)]}{R^\star_e},
    0,
    1
    \right),
\]
where $R_e(\pi)$ is the episodic return of policy $\pi$ in MuJoCo environment $e$, and $R^\star_e$ is an environment-specific reference return used only for normalization (see Table~\ref{tab:mujoco_friction_envs}). 
This quantity is not a binary success rate; rather, it is an SR-like competence proxy in $[0,1]$ used by archive insertion gates, anchor storage, few-shot selection, and revisit metrics. 
The raw MuJoCo return remains the fitness (reward) used for policy quality, while $\widetilde{\mathrm{SR}}$ provides a scale-compatible analogue of MiniGrid success rate.

\begin{table}[t]
\centering
\small
\setlength{\tabcolsep}{5pt}
\begin{tabular}{llccc}
\toprule
\textbf{Environment} & \textbf{Morphology / task} & \textbf{Friction scales} & \textbf{Visits} & \textbf{$R^\star_e$} \\
\midrule
\texttt{HalfCheetah-v5} & 2D runner / running & $\{0.50,0.75,1.00,1.25,1.50\}$ & $5 \times 2$ & 5000 \\
\texttt{Hopper-v5}      & 2D monoped / hopping & $\{0.50,0.75,1.00,1.25,1.50\}$ & $5 \times 2$ & 3000 \\
\texttt{Walker2d-v5}    & 2D biped / walking   & $\{0.50,0.75,1.00,1.25,1.50\}$ & $5 \times 2$ & 4000\\
\bottomrule
\end{tabular}
\caption{
\textbf{MuJoCo friction-shift curricula.}
Each MuJoCo environment is run as a separate same-morphology continual-learning problem. 
Within each environment, the agent cycles through five friction regimes twice, producing ten task visits. 
This preserves policy-parameter compatibility while inducing continuous-control dynamics shifts.
}
\label{tab:mujoco_friction_envs}
\end{table}

\paragraph{Methods.}
The primary method in this appendix is \textsc{TeLAPA} with the same core archive, few-shot transfer, and latent-maintenance mechanisms used in the MiniGrid experiments. 
At each task boundary, \textsc{TeLAPA} may select a transfer origin from previous archives using few-shot adaptation on the incoming friction regime. 
After training, MAP-Elites-style illumination samples local parameter-space perturbations around the trained policy, evaluates them under the current friction regime, embeds their rollout behavior, and stores competent yet behaviorally distinct policies in the archive. 
For comparison, we additionally run single-policy baselines in the same friction-shift setting: \textsc{Scratch}, \textsc{Scratch-Reuse}, \textsc{Finetune}, and \textsc{Finetune-Reset}. 
These baselines isolate whether gains come from archive-based transfer rather than merely from carrying forward the latest policy.

\begin{table}[t]
\centering
\small
\setlength{\tabcolsep}{5pt}
\begin{tabular}{lll}
\toprule
\textbf{Component} & \textbf{Hyperparameter} & \textbf{Value(s)} \\
\midrule
Curriculum 
& MuJoCo environments 
& \texttt{HalfCheetah-v5}, \texttt{Hopper-v5}, \texttt{Walker2d-v5}, \texttt{Ant-v5} \\
Curriculum 
& Friction scales 
& $\{0.50,\;0.75,\;1.00,\;1.25,\;1.50\}$ \\
Curriculum 
& Number of cycles 
& $2$ \\
Training 
& PPO steps per task visit 
& $1\times 10^6$ \\
Training 
& PPO update/evaluation chunk 
& $5\times 10^4$ \\
Training 
& Parallel training environments 
& $8$ \\
Evaluation 
& Evaluation episodes 
& $5$ \\
Evaluation 
& Parallel evaluation environments 
& $4$ \\
PPO 
& Policy architecture 
& MLP policy \\
PPO 
& Learning rate 
& $3\times 10^{-4}$ \\
PPO 
& Rollout length 
& $2048$ \\
PPO 
& Batch size 
& $256$ \\
PPO 
& Epochs per update 
& $10$ \\
PPO 
& Discount factor 
& $0.99$ \\
PPO 
& GAE $\lambda$ 
& $0.95$ \\
PPO 
& Clip range 
& $0.20$ \\
\textsc{TeLAPA} archive 
& MAP-Elites iterations 
& $200$ \\
\textsc{TeLAPA} archive 
& Candidate evaluation episodes 
& $5$ \\
\textsc{TeLAPA} archive 
& Mutation scale $\sigma$ 
& $0.02$ \\
\textsc{TeLAPA} archive 
& Archive competence score 
& $\widetilde{\mathrm{SR}}$ from normalized return \\
\textsc{TeLAPA} embedder 
& Trajectory feature dimension 
& $24$ \\
\textsc{TeLAPA} embedder 
& Latent dimension 
& $8$ \\
\bottomrule
\end{tabular}
\caption{
\textbf{MuJoCo \textsc{TeLAPA} hyperparameters.}
We use the same archive-based transfer structure as in MiniGrid and adapt only the environment interface, trajectory sketch, and dense-return competence normalization. 
}
\label{tab:mujoco_telapa_hparams}
\end{table}

\begin{table*}[ht]
\centering
\small
\setlength{\tabcolsep}{5.0pt}
\begin{tabular}{llccc}
\toprule
\textbf{Environment} & \textbf{Method} 
& \textbf{Mean $\widetilde{\mathrm{SR}}$ $\uparrow$} 
& \textbf{BWT $\uparrow$} 
& \textbf{TR $\uparrow$} \\
\midrule
\multirow{5}{*}{\texttt{HalfCheetah-v5}} & \textsc{Scratch} & $0.234 \pm 0.003$ & $-0.002 \pm 0.015$ & $0.000 \pm 0.000$ \\
 & \textsc{Scratch-Reuse} & $0.292 \pm 0.006$ & $0.038 \pm 0.033$ & $0.228 \pm 0.006$ \\
 & \textsc{Finetune} & $\mathbf{0.379 \pm 0.029}$ & $0.036 \pm 0.009$ & $\mathbf{0.396 \pm 0.031}$ \\
 & \textsc{Finetune-Reset} & $0.376 \pm 0.026$ & $0.031 \pm 0.007$ & $0.393 \pm 0.026$ \\
 & \textsc{TeLAPA} & $0.317 \pm 0.005$ & $\mathbf{0.051 \pm 0.004}$ & $0.298 \pm 0.006$ \\
\midrule
\multirow{5}{*}{\texttt{Hopper-v5}} & \textsc{Scratch} & $0.618 \pm 0.014$ & $-0.540 \pm 0.015$ & $-0.004 \pm 0.003$ \\
 & \textsc{Scratch-Reuse} & $0.620 \pm 0.016$ & $-0.612 \pm 0.010$ & $0.576 \pm 0.024$ \\
 & \textsc{Finetune} & $0.603 \pm 0.014$ & $-0.581 \pm 0.024$ & $0.394 \pm 0.037$ \\
 & \textsc{Finetune-Reset} & $0.593 \pm 0.023$ & $-0.551 \pm 0.013$ & $0.352 \pm 0.034$ \\
 & \textsc{TeLAPA} & $\mathbf{0.629 \pm 0.041}$ & $\mathbf{-0.179 \pm 0.060}$ &  $\mathbf{0.740 \pm 0.038}$ \\
\midrule
\multirow{5}{*}{\texttt{Walker2d-v5}} & \textsc{Scratch} & $0.264 \pm 0.014$ & $-0.094 \pm 0.030$ & $-0.002 \pm 0.002$ \\
 & \textsc{Scratch-Reuse} & $0.476 \pm 0.024$ & $-0.212 \pm 0.051$ & $0.243 \pm 0.018$ \\
 & \textsc{Finetune} & $\mathbf{0.883 \pm 0.011}$ & $-0.395 \pm 0.041$ & $0.631 \pm 0.025$ \\
 & \textsc{Finetune-Reset} & $0.872 \pm 0.010$ & $-0.467 \pm 0.083$ & $0.653 \pm 0.039$ \\
 & \textsc{TeLAPA} & $0.779 \pm 0.035$ & $\mathbf{0.215 \pm 0.038}$ & $\mathbf{0.763 \pm 0.020}$ \\
\bottomrule
\end{tabular}
\caption{
\textbf{Continuous-control friction-shift results.} Metrics are computed over the two-cycle, five-friction MuJoCo curriculum for each environment. $\widetilde{\mathrm{SR}}$ denotes normalized return competence, BWT measures retention across revisits, and TR measures transfer recovery.
}
\label{tab:mujoco_friction_results}
\end{table*}

\begin{figure}[ht]
  \centering
  \includegraphics[width=\linewidth]{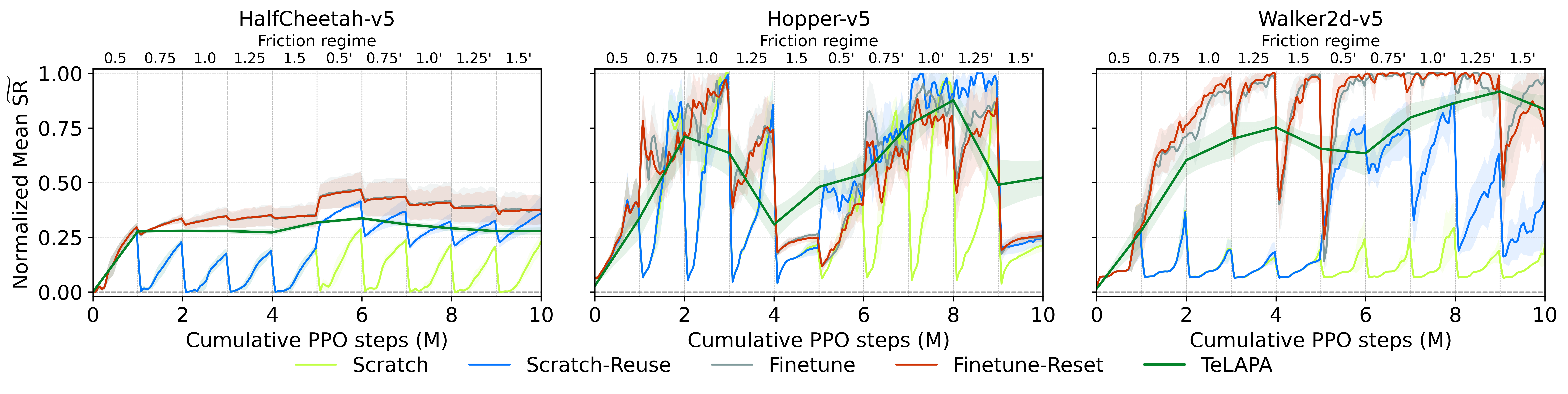}
  \caption{
    \textbf{Learning curves under continuous-control friction shifts.}
    We evaluate each method on a two-cycle MuJoCo curriculum with five friction regimes per environment.
    The x-axis reports cumulative PPO training steps, while the y-axis reports normalized return competence $\widetilde{\mathrm{SR}}$.
    Vertical dotted lines mark friction-regime transitions, and top-axis labels indicate the active friction regime, with primes denoting revisits in the second cycle.
    Compared with \textsc{Scratch} and \textsc{Scratch-Reuse}, which often collapse after each dynamics shift, \textsc{TeLAPA} maintains smoother competence across regime boundaries.
    Compared with finetuning variants, \textsc{TeLAPA} does not always reach the highest peak competence, but it exhibits stronger retention and transfer recovery, consistent with the BWT and TR trends in Table~\ref{tab:mujoco_friction_results}.
    Shaded regions indicate variability across runs.
    }
\label{fig:mujoco_friction_curves}
\end{figure}

\paragraph{Results.}
Table~\ref{tab:mujoco_friction_results} and Fig.~\ref{fig:mujoco_friction_curves} show that \textsc{TeLAPA} transfers its archive-based recovery mechanism to continuous-control dynamics shifts, although its advantage differs across environments.
In \texttt{Hopper-v5}, \textsc{TeLAPA} obtains the best mean normalized competence, BWT, and TR, indicating that its archived policy neighborhoods substantially improve recovery under repeated friction changes.
In \texttt{Walker2d-v5}, \textsc{Finetune} and \textsc{Finetune-Reset} methods reach higher peak competence, but \textsc{TeLAPA} achieves the strongest BWT and TR, suggesting better retention and reuse across revisits even when it does not maximize final per-task performance.
In \texttt{HalfCheetah-v5}, \textsc{Finetune} attains the highest mean competence, while \textsc{TeLAPA} obtains the best BWT, again indicating a retention advantage rather than a pure peak-performance advantage.
The learning curves further show that \textsc{TeLAPA} produces smoother adaptation across friction boundaries.
Unlike \textsc{Scratch} and \textsc{Scratch-Reuse} which repeatedly collapse after each regime change, and unlike \textsc{Finetune} and \textsc{Finetune-Reset}, which often achieve high competence but exhibit sharper drops at transitions, \textsc{TeLAPA} maintains a more stable competence trajectory across the two-cycle curriculum.
Thus, the MuJoCo ablation supports the same qualitative interpretation as the MiniGrid results: \emph{preserving a set of reusable source-competent policy neighborhoods improves recoverability under non-stationary task structure, even when the method does not always achieve the highest performance score.}

\paragraph{Limitations and scope.}
This MuJoCo experimental ablation is intended as a proof of concept rather than a complete continuous-control benchmark. 
First, each MuJoCo curriculum keeps the morphology fixed and only varies friction. 
Thus, the experiment tests transfer across dynamics shifts but does not test transfer across incompatible observation/action spaces such as \texttt{HalfCheetah}$\rightarrow$\texttt{Hopper}. 
We designed this experiment with this constraint in mind. 
\begin{enumerate}
    \item Our current version of \textsc{TeLAPA} archives full policy parameter vectors. 
    Designing and implementing a fully-fledge cross-morphology transfer would require additional mechanisms for architecture and morphology policy translation. 
    \item MuJoCo provides continuous scalar rewards instead of binary success signals.
    Therefore, we report normalized return competence $\widetilde{\mathrm{SR}}$ as an SR-like proxy for thresholding and retention metrics. 
    \item The MuJoCo trajectory descriptor is a compact continuous-control proof-of-concept rather than a more exact behavioral descriptor. 
    While compatible with our MiniGrid latent-maintenance algorithm, we acknowledge that it is not meant to be the final or most expressive descriptor for continuous-control behavior.
    \item MuJoCo and MiniGrid have different complexities, however, our experiments used the same TeLAPA optimal values found in Table~/\ref{tab:hp_search_telapa}. 
    We did not perform any sweep over hyperparameters.
    Thus, our results are sub-optimal. 
    Nonetheless, the results still show the capabilities of \textsc{TeLAPA}'s main ideas.
\end{enumerate}

Despite these limitations, the experiment addresses two central scope concerns: 
(i) \textsc{TeLAPA}'s core mechanism is not tied to image observations, symbolic missions, or sparse MiniGrid success events; and 
(ii) the same archive, few-shot origin selection, MAP-Elites illumination, and latent-space maintenance algorithm can be applied to continuous-control dynamics environments.
Thus, this experiment serves as a controlled test to show that the ideas introduced and explored by \textsc{TeLAPA} such as preserving and retrieving a neighborhood of competent policies can support recoverable adaptation under non-stationarity beyond the original MiniGrid setting.

\end{document}